\documentclass[acmsmall]{acmart}
\renewcommand\footnotetextcopyrightpermission[1]{}
\usepackage{enumitem}
\usepackage[capitalise]{cleveref}
\usepackage{times}
\usepackage{epsfig}
\usepackage{graphicx}
\usepackage{amsmath}
  
\usepackage{amssymb}
\usepackage{array}
\usepackage{tabularx}
\usepackage{threeparttable}
\usepackage{multirow}
\usepackage{makecell}
\usepackage{colortbl}
\usepackage{hhline}
\usepackage{booktabs}
\AtBeginDocument{%
  }

\setcopyright{acmlicensed}
\copyrightyear{2024}
\acmYear{2024}
\acmDOI{XXXXXXX.XXXXXXX}

\acmISBN{978-1-4503-XXXX-X/18/06}

\begin{document}

\title{Vertical Federated Learning for Effectiveness, Security, Applicability: A Survey}

\author{Mang Ye}
\email{yemang@whu.edu.cn}
\author{Wei Shen}
\email{weishen@whu.edu.cn}
\author{Bo Du}
\email{dubo@whu.edu.cn}
\affiliation{%
  \institution{The National Engineering Research Center for Multimedia Software, School of Computer Science, Wuhan University}
  \city{Wuhan}
  \country{China}
}

\author{Eduard Snezhko}
\email{eduard.snezhko@gmail.com}
\author{Vassili Kovalev}
\email{vassili.kovalev@gmail.com}
\affiliation{%
  \institution{United Institute of Informatics Problems of Belarus National Academy of Sciences, National Academy of Sciences of Belarus}
  \city{Minsk}
  \country{Belarus}
}

\author{Pong C. Yuen}
\email{pcyuen@comp.hkbu.edu.hk}
\affiliation{%
  \institution{Department of Computer Science, Hong Kong Baptist University}
  \city{Hong Kong}
  \country{China}
}

\renewcommand{\shortauthors}{Ye et al.}

\begin{abstract}
  Vertical Federated Learning (VFL) is a privacy-preserving distributed learning paradigm where different parties collaboratively learn models using partitioned features of shared samples, without leaking private data. Recent research has shown promising results addressing various challenges in VFL, highlighting its potential for practical applications in cross-domain collaboration. However, the corresponding research is scattered and lacks organization. To advance VFL research, this survey offers a systematic overview of recent developments. First, we provide a history and background introduction, along with a summary of the general training protocol of VFL. We then revisit the taxonomy in recent reviews and analyze limitations in-depth. For a comprehensive and structured discussion, we synthesize recent research from three fundamental perspectives: effectiveness, security, and applicability. Finally, we discuss several critical future research directions in VFL, which will facilitate the developments in this field. We provide a collection of research lists and periodically update them at \href{https://github.com/shentt67/VFL_Survey}{https://github.com/shentt67/VFL\_Survey}.
\end{abstract}





\maketitle

\section{Introduction}
The rapid development of deep learning has created a significant demand for high-quality and large-quantity data. However, collecting sufficient data for training within a single party is often challenging. An intuitive solution is to collaborate by sharing data among multiple associated participants, but this valuable data often involves personal privacy concerns \cite{fung2010privacy, wagner2018technical, liu2021machine, salehzadeh2024wearable, ye2024securereid, han2023privacy} or confidentiality agreements \cite{may2005intellectual, international2016toward, gdpr2018general, ccpa2018, cdpa2018}, which hinders data sharing and leads to data silos. As an alternative solution, Federated Learning (FL) \cite{yang2019federated, mcmahan2017communication, li2020federated, yin2021comprehensive, huang2023federated, ye2023heterogeneous} has garnered increasing attention in research and applications. FL aims to collaboratively train models across participants without exposing raw data. Based on the distributed way of data, Federated Learning can be primarily categorized into three scenarios: 

\begin{itemize}[leftmargin=*]
  \item \textbf{Horizontal Federated Learning (HFL)} \cite{yang2019federated, mcmahan2017communication, li2020federated, yin2021comprehensive, nguyen2022federated, pfeiffer2023federated, wu2023topology, qu2022blockchain, fang2022robust, huang2022learn, huang2023federated, huang2023federated_survey, ye2023heterogeneous, huang2023rethinking, fang2023robust, yang2023dynamic, huang2023generalizable, shang2022federated, huang2024federated, FedHEAL_CVPR2024}: Here, participants share the same feature space but have different local samples. The goal is to train a global model that can generalize across samples from different clients.

  \item \textbf{Vertical Federated Learning (VFL)} \cite{hardy2017private, yang2019federated, liu2024vertical}: In this scenario, clients share the same samples (shared/aligned samples) but have different local features. The goal of VFL is to train a global model capable of making predictions using the distributed features of shared samples.

  \item \textbf{Federated Transfer Learning (FTL)} \cite{yang2019federated, saha2021federated, liu2020secure, chen2020fedhealth}: Additionally, there is a scenario where clients share both common samples and parts of the feature spaces. The goal of FTL is to transfer knowledge from one client to others through shared samples.
\end{itemize}

As a specific scenario of federated learning, VFL finds extensive application in cross-domain scenarios. For example, as illustrated in~\cref{fig:example_VFL}, a mall could collaborate with a video platform and a bank to train a global model that predicts shared users' shopping intentions. Typically, participants in VFL include an active client and several passive clients. Each client transforms raw sample features into feature embeddings using local models. These are then sent to the active client and aggregated by a global model to make predictions. Additionally, a trustworthy coordinator is employed to ensure secure communication and sample alignment~\cite{hardy2017private, liang2004privacy, lu2020multi, benny2014faster, meadows1986more, pinkas2015phasing, chen2017fast, pinkas2018scalable}. This setup facilitates cross-domain collaboration without compromising data privacy. Recently, VFL has been widely explored and shown promising results in various domains, including finance \cite{cheng2020federated, zheng2020vertical, long2020federated}, recommendation systems \cite{zhang2021vertical, wei2023fedads, yuan2022privacy, cai2020bytedance, lin2021practice, wei2023fedads}, and healthcare \cite{yan2024cross, huang2023vertical, hu2022vertical, rooijakkers2020convinced, song2021federated}, among others \cite{li2020review, liu2021fate, he2020fedml, teimoori2022secure, niknam2020federated, zhang2020vertical, hashemi2021vertical, liu2021federated, ge2022failure}.

Despite the development and application of VFL, various challenges still hinder its progress. Researchers have explored these challenges, proposing feasible solutions across three main areas: effectiveness, security, and applicability. \textit{Effectiveness} focuses on improving VFL under typical conditions, including the development of superior models and the selection of optimal features and clients; \textit{Security} is vital, as VFL involves collaboration among multiple parties. Ensuring data safety and protecting against attacks is essential to maintaining trust among all participants; \textit{Applicability} assesses how VFL can operate effectively in real-world scenarios, which are often constrained. This exploration helps to broaden the practical utility of VFL across different sectors. Together, these themes construct a comprehensive research landscape:

\begin{figure}
    \centering
    \includegraphics[width=0.85\linewidth]{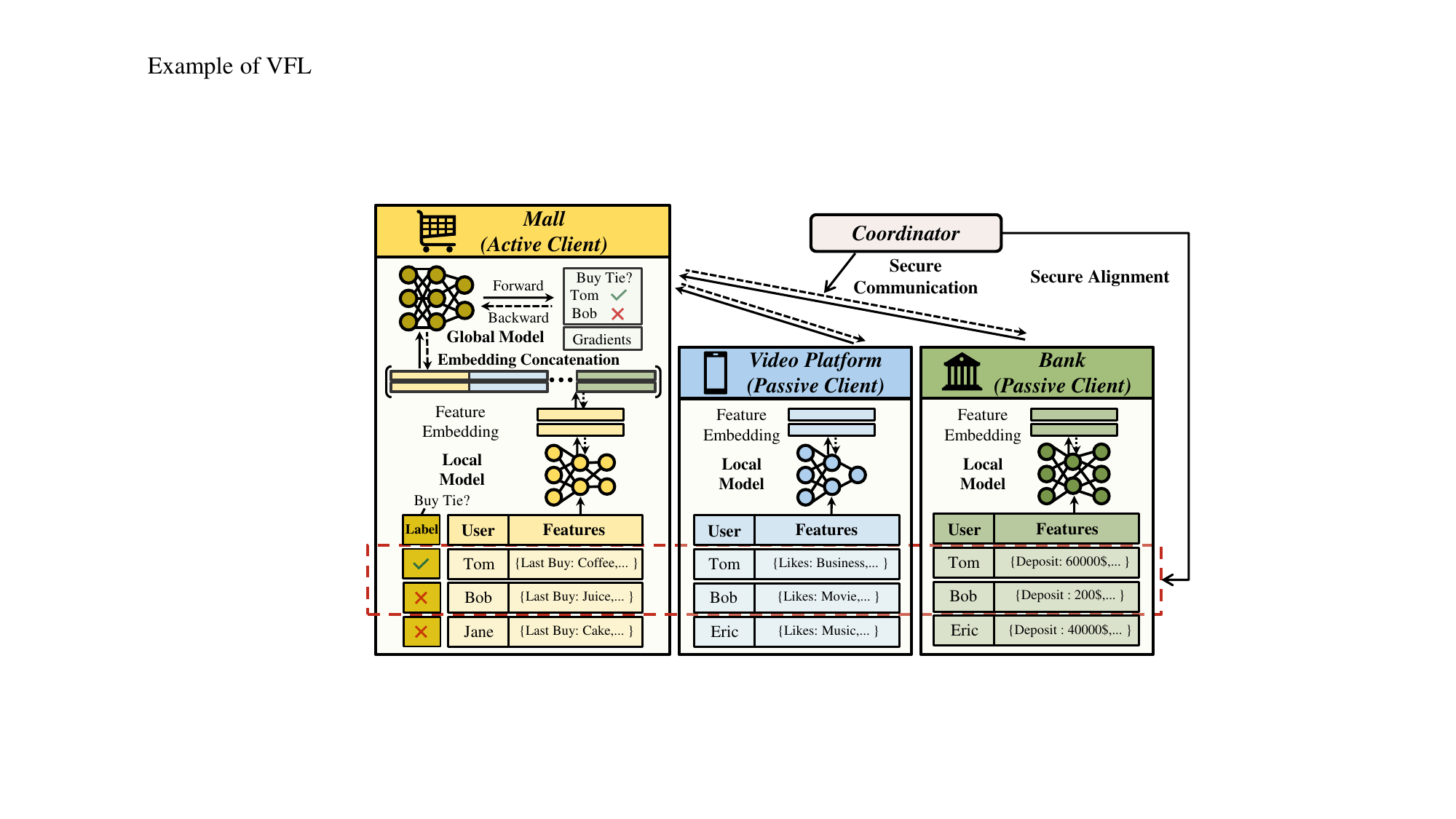}
    \caption{\textbf{A Practical Application of Vertical Federated Learning.} We present a practical cross-domain collaboration with three participants: mall, video platform, and bank. The mall acts as the active client, collaborating with the video platform and the bank as passive clients. Each client holds the local features and models of the same users. The active client holds the task labels, e.g., whether to buy the tie. A global model is introduced to make the final prediction of the shared/aligned users by aggregating feature embeddings from all clients. With prediction results and labels, the gradients can be calculated for both global and local model updation. Besides, a third-party coordinator can be employed for secure communication and sample alignment. }
    \label{fig:example_VFL}
\end{figure} 

\begin{itemize}[leftmargin=*]
\item \textbf{Effectiveness.} In machine learning, it is significant to mine effective information from statistical data, which promotes the target learning tasks \cite{mitchell1999machine, al2015efficient}. In VFL, when collaborating with splitting features across multiple participants, the similar and crucial challenge is to aggregate effective information from features in a distributed way. For the general setting in VFL, there are two research directions to achieve effective distributed feature aggregation and task learning: (1) \textit{Model Design.} These works aim to design basic architecture that adapts for distributed feature learning. Specifically, designing effective models for collaboration with distributed features is the foundation of constructing the VFL paradigm in practical cross-domain collaboration. The topics include extracting meaningful embeddings with split features, achieving effective global feature aggregation, privacy-preserving learning models, and so on. Generally, there are two different kinds of models: (i) Tree-based model. Tree-based model is light and highly efficient, which is suitable for simple data structures such as tabular data. (ii) Neural network-based model. Compared with the traditional machine learning paradigm, the Neural network-based model has enhanced expression ability, which is more suitable for complex multi-modal data, such as text, image, and audio. (2) \textit{Feature \& Client Selection.} In VFL, each client has a contribution to the prediction of shared/aligned samples. However, due to the client heterogeneity, the influences on the final prediction are different between clients. Concretely, the information contained in different features is various, and the local model architectures are different. which affects the results of the prediction task. In this case, recent research explores selecting critical features or clients to obtain effective information from raw data, promoting advanced collaboration performance. The core idea of feature \& client selection is to adjust the contribution of different features and clients, which eliminates interference from other redundant information.

\begin{figure}
    \centering
    \includegraphics[width=0.95\linewidth]{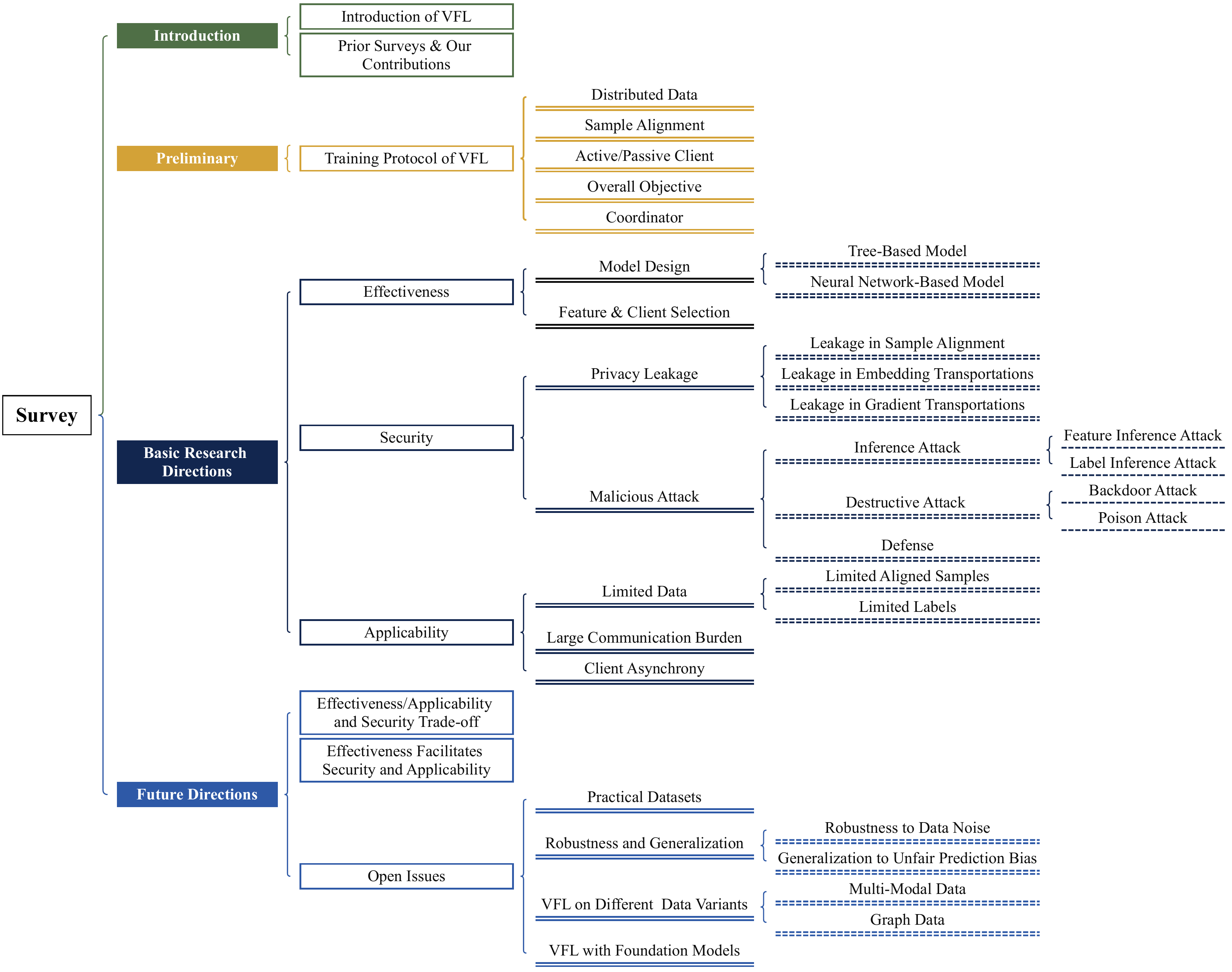}
    \caption{\textbf{Survey Outline.} It concludes four parts: Introduction, Preliminary, Basic Research Directions, and Future Directions.}
    \label{fig:Survey_Outline}
\end{figure}

\item \textbf{Security.} Vertical Federated Learning is proposed as a privacy-preserving paradigm for distributed features. Although primary privacy ensurance is achieved without leaking raw data, enhanced security is needed to handle the different security threats. In summary, there are two aspects of security concerns: (1) \textit{Privacy Leakage.} Firstly, privacy leakage problems exist in different stages of VFL, including leakage in sample alignment, embedding transportations, and gradient transportations. (2) \textit{Malicious Attack.} On the one hand, to obtain the raw data information, the attackers perform data inference attacks: Firstly, all clients can perform the feature inference attack to get the original features in other clients. Besides, the passive clients can perform label inference attacks to infer the labels of the samples from active clients. Moreover, the attackers can perform destructive attacks to affect the system: The attackers can implant specific backdoors to the target prediction, or inject data perturbation to perform poison attack for degrading the collaboration performance. It is also crucial to design corresponding defense strategies against different attacks. 

\item \textbf{Applicability.} When deploying vertical federated learning in various practical scenarios, different challenges hinder the performance of VFL. It is critical to promote practical application in different limited situations. When deploying VFL in practical scenarios, there exists the following challenges: (1) \textit{Limited Data.} In VFL, there are two aspects of limited data, that hinder the generalization in practical scenarios: (i) Limited Aligned Samples. The critical step to deploying VFL is to find the shared/aligned samples between clients. However, the shared clients are usually in few amounts, especially with a large number of clients. In the situation of limited aligned samples, the performance of VFL is trapped by the limited knowledge and poor generalization of the few aligned samples. (ii) Limited Labels. The labels of the active client are the crucial guidance for collaboration task learning. However, due to various reasons such as the difficulty in collecting large quantities or high-quality labels, the amount of accessible labels for collaboration is often limited, which hinders the performance. (2) \textit{Large Communication Burdens.} To achieve collaboration across clients, multiple communications are required, which brings large communication burdens to the practical systems. At the same time, the resources in the system are often limited. In this case, it is difficult to apply VFL in practical systems. It is critical to design a communication-effective framework to increase the applicability of VFL in practical scenarios.  (3) \textit{Client Asynchrony.} For the differences in models, number of features, calculation ability, and so on, the clients will finish their calculation and communication asynchronously in a single training epoch, which causes a huge updation delay and training time costs.
\end{itemize}

As Vertical Federated Learning (VFL) gains significance in practical applications and emerges as a critical research field, there are several recent surveys \cite{yang2019federated, xu2022privacy,khan2022vertical, wei2022vertical, li2023vertical, yang2023survey, liu2024vertical, yu2024survey} that explore general research topics within VFL. However, the landscape of existing research is fragmented and complex, with no comprehensive and rational taxonomy of research challenges in VFL that encompasses the most recent methods. Yang \textit{et al.} \cite{yang2019federated} offer a pioneering systemic definition of Federated Learning categorized by data distribution, discussing potential applications. However, it is a general survey on federated learning (FL), lacking a specific and thorough investigation of vertical federated learning (VFL). Liu \textit{et al.} \cite{liu2024vertical} conduct a literature survey on the aspects of effectiveness, efficiency, and privacy, with a unified framework for VFL. However, with the rapid initiation and development of the field, some of the latest literature, including aspects of applicability and security, have not been included. Besides, our survey provides an additional discussion and summary of future research directions compared to this work. While other surveys \cite{xu2022privacy, khan2022vertical, wei2022vertical, li2023vertical, yang2023survey, yu2024survey} summarize existing works focusing on scattered aspects such as privacy concerns \cite{xu2022privacy, li2023vertical, yu2024survey}, experimental settings \cite{khan2022vertical, wei2022vertical, li2023vertical}, database systems \cite{yang2023survey}, and theoretical analysis \cite{li2023vertical, yang2023survey}, with fragmented summaries of existing works, we provide a summary from three fundamental perspectives: effectiveness, security, and applicability, which encompasses the most recent works and offers a comprehensive landscape.

In conclusion, we propose a novel taxonomy method aimed at providing an exhaustive survey that incorporates the latest advances. We categorize existing works into three aspects, which comprehensively summarize the most recent developments: Effectiveness, which ensures the performance of VFL in general settings; Security, which provides advanced guarantees against privacy concerns and malicious attackers; and Applicability, which explores solutions for developing VFL algorithms in various practical applications. Based on our summary, we propose a series of future research topics that are both unsolved and valuable. The outline of this survey is illustrated in \cref{fig:Survey_Outline}. The principal contributions of this survey can be outlined as follows:

\begin{itemize}[leftmargin=*]
\item We provide an in-depth analysis of vertical federated learning and present the state-of-the-art comprehensive survey on its effectiveness in general VFL scenarios, security concerning privacy issues and malicious attacks, and applicability in limited practical applications.
\item We collect updated and influential works from prominent conferences and journals, as well as recent literature on arXiv, and organize them into a novel taxonomy: Model Design and Feature \& Client Selection under Effectiveness, Privacy Leakage and Malicious Attacks under Security, and Limited Data, Large Communication Burden, and Client Asynchrony under Applicability. In-depth analyses and comparisons of these methods are also included.
\item We provide an extensive discussion of future directions, including the trade-offs and synergies between basic directions and other open issues, which will serve as references for potential research and applications in this field.
\end{itemize}

\section{Preliminary}
In this section, we will describe the general setting of VFL, which is also illustrated in recent literature \cite{yang2019federated, liu2024vertical, xu2022privacy, khan2022vertical, wei2022vertical, li2023vertical, yang2023survey}. Compared with existing works, we will provide a more comprehensive summary as follows:

\textbf{Distributed Data.} In VFL, the data is distributed across clients by feature spaces. Suppose there are $N$ data samples in total, and the number of clients is $M$. The sample set is defined as $D=\left \{ x_{i}, y_{i} \right \}_{i=1}^{N}$, where $x_{i} \in \mathbb{R}^{d}$ is the dimension number of raw features, $y_{i}$ is the corresponding label. The distributed data has two situations: if client $k$ is the active client, then the data is defined as $D^{k}=\left \{ x_{i}^{k}, y_{i}^{k} \right \}_{i=1}^{N^{k}}, x_{i} \in \mathbb{R}^{d^{k}}$ where $y_{i}^{k}$ is the corresponding labels of the sample $x_{i}^{k}$; if client $k$ is the passive client, the data can be defined as $D^{k}=\left \{ x_{i}^{k} \right \}_{i=1}^{N^{k}}, x_{i} \in \mathbb{R}^{d^{k}}$. In above definition, the $N^{k}$ represents the number of the samples in client $k$, and $d^{k}$ is the number of feature dimensions in client $k$, which satisfies $d=\sum^{M-1}_{k=1}d^{k}$. 

\textbf{Sample Alignment.} Denote the shared sample set as $D^{S}=\left \{ x_{i}^{S}, y_{i}^{S} \right \}_{i=1}^{N^{S}}, x_{i} \in \mathbb{R}^{d}$, where $y_{i}^{S}$ is the corresponding labels and origin from the active client. The clients in VFL share the same samples with distributed features, and the shared/aligned samples are discovered with the Sample Alignment process. There are two kinds of sample alignment methods explored in recent research~\cite{liang2004privacy, lu2020multi, benny2014faster, meadows1986more, pinkas2015phasing, chen2017fast, pinkas2018scalable, sun2021vertical, kissner2005privacy, davidson2017efficient, jia2022shuffle, inbar2018efficient, kolesnikov2017practical, hardy2017private}: One of the ideal solutions is joint the data across clients by the common unique IDs for samples; Besides, the raw data in different clients can be encrypted and sent to the trust-worthy third party, pairing by matching strategy on encrypted data. 

\textbf{Active/Passive Client.} There are two different kinds of clients in VFL, representing different statuses in the collaboration process. Generally, there is one active client in a single VFL process. The active client is the initiator of collaboration, where the labels of the collaboration task are stored. Denote the set of labels as $Y=\left \{y_{i} \right \}_{i=1}^{N}$, and $y_{i}$ represents the corresponding labels of the sample $x_{i}$. Besides, the passive clients hold parts of sample features only, and all the features from both active and passive clients are unioned by sample alignment for model training, under the supervision of corresponding labels from the active client. 

\begin{figure}
    \centering
    \includegraphics[width=1\linewidth]{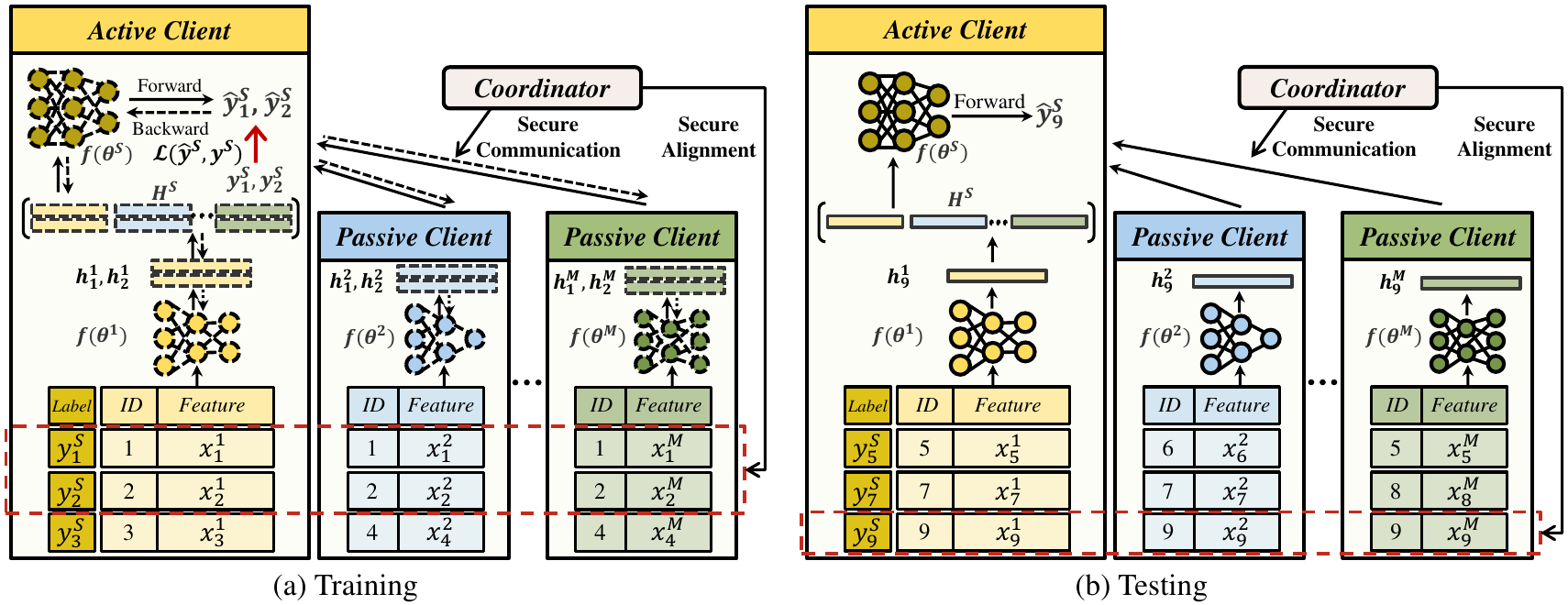}
    \caption{\textbf{The general flow of training and testing in VFL.} (a) During training, aligned sample embeddings are sent to the active client, where gradients are calculated based on task labels. The overall objective is to optimize for collaborative prediction. These gradients are then sent back to each client for model updating. (b) During testing, predictions on aligned samples are made utilizing the trained global and local models.}
    \label{fig:Train_Test}
\end{figure}

\textbf{Overall Objective.} The goal of VFL is to train a global model for accurate prediction on the shared samples $D^{S}=\left \{ x_{i}^{S}, y_{i}^{S} \right \}_{i=1}^{N^{S}}$. Denote the local model in client $k$ as $F(\theta^{k})$, which is employed for learning embeddings from the origin local data. Firstly, the local data $x_{i}^{k}$ in client $k$ is transformed into the feature embedding $h_{i}^{k}$ by local model, which is defined as follows:

\begin{equation}
\setlength{\abovedisplayskip}{0pt}
\setlength{\belowdisplayskip}{0pt}
{
    {h}^{k}_{i} = F(x^{k}_{i};\theta^{k}).
}
\label{eq:embedding}
\end{equation}

After that, the embedding concatenation of shared samples $H^{S}= [h^{k}], k=1,...,M-1$ are transmitted to the side of the global model $G(\theta^{S})$. The global model can be stored in the active client, or a trust-worthy server. Then, the embeddings are aggregated to predict labels of shared samples. The predicted labels $\hat y^{S}$ for the samples $x^{S} \in D^{S}$ are defined as:

\begin{equation}
\setlength{\abovedisplayskip}{0pt}
\setlength{\belowdisplayskip}{0pt}
{
    \hat y^{S} = G(H^{S}; \theta^{S}).
}
\label{eq:label_prediction}
\end{equation}

The objective of collaborative learning is defined as follows:

\begin{equation}
\setlength{\abovedisplayskip}{0pt}
\setlength{\belowdisplayskip}{0pt}
{
    \theta^{1},...,\theta^{k}, \theta^{S} = \arg \min_{\theta^{1},...,\theta^{k}, \theta^{S}}\mathcal{L}(\hat y^{S},y^{S}),
}
\label{eq:optimization}
\end{equation}
where $\mathcal{L}$ is the loss function of the collaboration learning task. For instance, an alternative loss function could be the cross-entropy loss for the classification task.  

\textbf{Coordinator.} To provide secure ensurance, a trust-worthy coordinator is introduced. It controls the privacy protocol of secure communication and the sample alignment process. In VFL, the coordinator can be the active client or a third-party server. The general flow of training and testing in VFL is illustrated in \cref{fig:Train_Test}. In the training stage, the local data is transformed into embeddings and aggregated by the global model, and then the gradients are backward to both the global model and local models for model updation. In the testing stage, the data of test samples is transformed and transmitted to the trained global model for prediction.

\section{Effectiveness}
In the typical setting of Vertical Federated Learning (VFL), data is distributed across different clients by feature, with the objective of training a global model within a privacy-preserving framework. The cornerstone of VFL lies in crafting a foundational paradigm for efficiently extracting information from these distributed features. To enhance effectiveness in the basic setup of VFL, efforts have concentrated on two main areas: Firstly, recent studies have focused on developing fundamental models for VFL that facilitate effective collaboration while ensuring privacy. Secondly, pioneering research has explored ways to extract more valuable information from all features or clients by assessing the importance of each feature or client. We will elaborate on these aspects in detail next.

\begin{figure}
    \centering
    \includegraphics[width=0.9\linewidth]{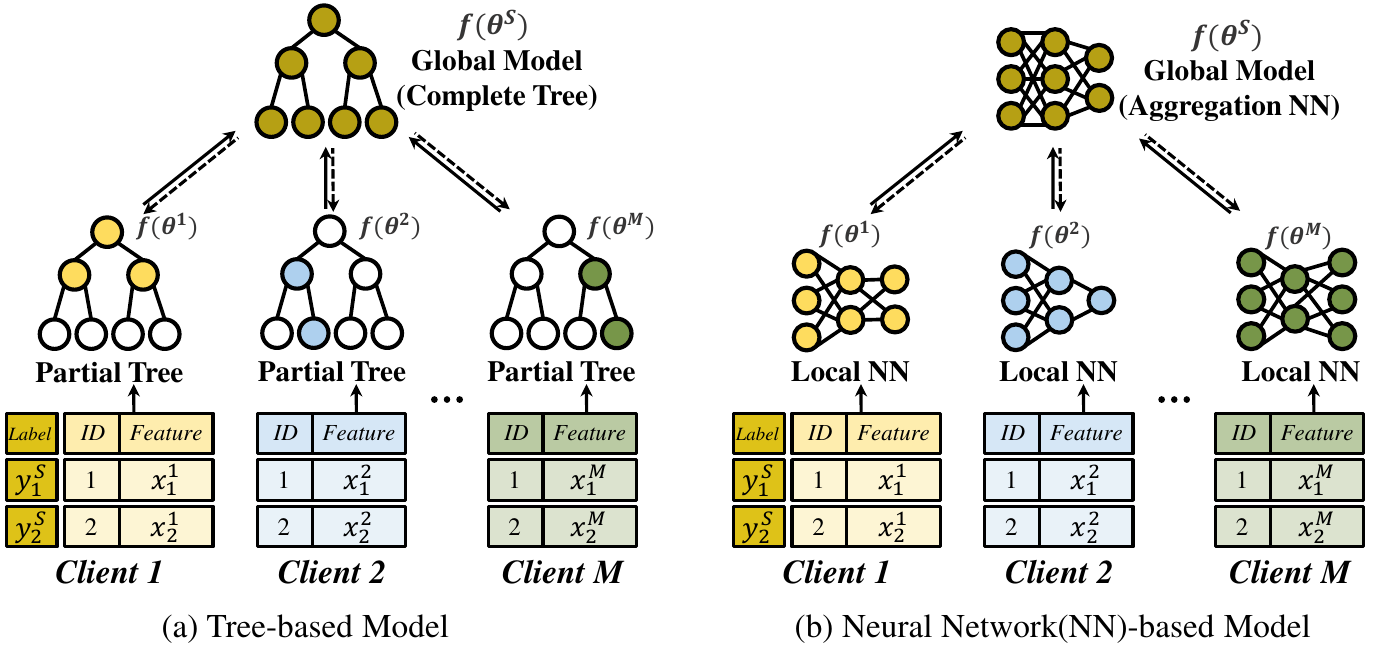}
    \caption{\textbf{The illustration of Tree-based/Neural Network-based model.} (a) For the tree-based model in VFL, each client trains a part of the tree model using partitioned features. These partial trees are then combined to construct the global tree model. (b) For the Neural Network-based model, each client trains its local model to extract embeddings. The global model is then trained in the active client using embeddings from all clients.}
    \label{fig:Model_Design}
\end{figure}

\subsection{Model Design}
In traditional machine learning, the raw data is centralized, and the whole model can be applied to learning the data in a single client. However, the traditional models are not applicable in VFL due to the decentralized features. The specific model designs are needed in VFL to ensure that decentralized features are aggregated in a privacy-preserving paradigm, which forms the foundation for effective collaboration. It can be categorized into two different types based on their basic architecture:

\textbf{Tree-based Model.} It is a lightweight design for distributed feature aggregation, suitable for making rapid predictions with modest accuracy. To maintain a basic privacy-preserving flow in VFL, several studies \cite{wu2020privacy, cheng2021secureboost, chen2021secureboost+, feng2019securegbm, fang2021large} propose utilizing encryption strategies with ID and feature embeddings to enable private alignment and aggregation, effectively training a split tree model. Other works \cite{tian2020federboost, li2022opboost} focus on constructing a privacy-preserving paradigm while achieving effective feature aggregation. Additionally, random forest-based approaches \cite{liu2020federated, yao2022efficient, hou2021verifiable} have been introduced, employing bagging strategies to enhance training effectiveness. Liu \textit{et al.} \cite{liu2020federated} suggest using the CART tree combined with bagging and a trustworthy third party to develop a random forest algorithm specifically for vertical federated learning. Yao \textit{et al.} \cite{yao2022efficient} utilize a random affine code to facilitate privacy-preserving collaboration, suitable for large-scale data applications. Hou \textit{et al.} \cite{hou2021verifiable} introduce a verifiable strategy for random forests in VFL to ensure effective collaboration among dynamically changing participants. Squirrel \cite{lu2023squirrel} enables two data owners to train a GBDT model on vertically split datasets without revealing sensitive information, safeguarding privacy against semi-honest adversaries. This framework combines novel GBDT algorithm designs with advanced cryptography, featuring efficient mechanisms for hiding sample distribution and optimized methods for gradient aggregation. However, it acknowledges the challenges of designing a secure and computationally efficient GBDT training protocol that is also communication-efficient under practical constraints such as limited high-speed bandwidth for cross-enterprise collaborative learning. The complexity of maintaining privacy during training and potential risks associated with publishing decision trees are highlighted as areas for further research. HybridTree \cite{li2023effective} incorporates party-specific knowledge by adding layers to the tree structure, enabling significant speed improvements in training times and providing a scalable solution for practical federated learning scenarios. However, the method is particularly suited for tabular data and may not directly apply to other data types like images or text.

\textbf{Neural Network-Based Model.} It leverages deep neural networks, well-suited for handling complex data types (such as images, text, and audio) and tasks (like detection and recognition) within the VFL framework. Romanini \textit{et al.} \cite{romanini2021pyvertical} introduce PyVertical, a foundational design for vertical federated learning that utilizes split neural networks. This allows participants to train on vertically partitioned data features across different clients while keeping their raw data private. Feng \textit{et al.} \cite{feng2020multi} explore a privacy-preserving label-sharing strategy. The proposed framework, MMVFL, facilitates effective collaboration among multiple participants through sparse learning and optimization with a global pseudo-label matrix. However, the assumption that data from all participants share the same label space limits the applicability of MMVFL. Liu \textit{et al.} \cite{liu2020secure} propose the Federated Transfer Learning (FTL) framework, which facilitates knowledge transfer from the source domain using overlapped samples and features. Addressing the gap for sequential data, Abedi \textit{et al.} \cite{abedi2024fedsl} introduce FedSL, a split Recurrent Neural Network tailored for distributed sequential data processing.

\begin{figure}
    \centering
    \includegraphics[width=1\linewidth]{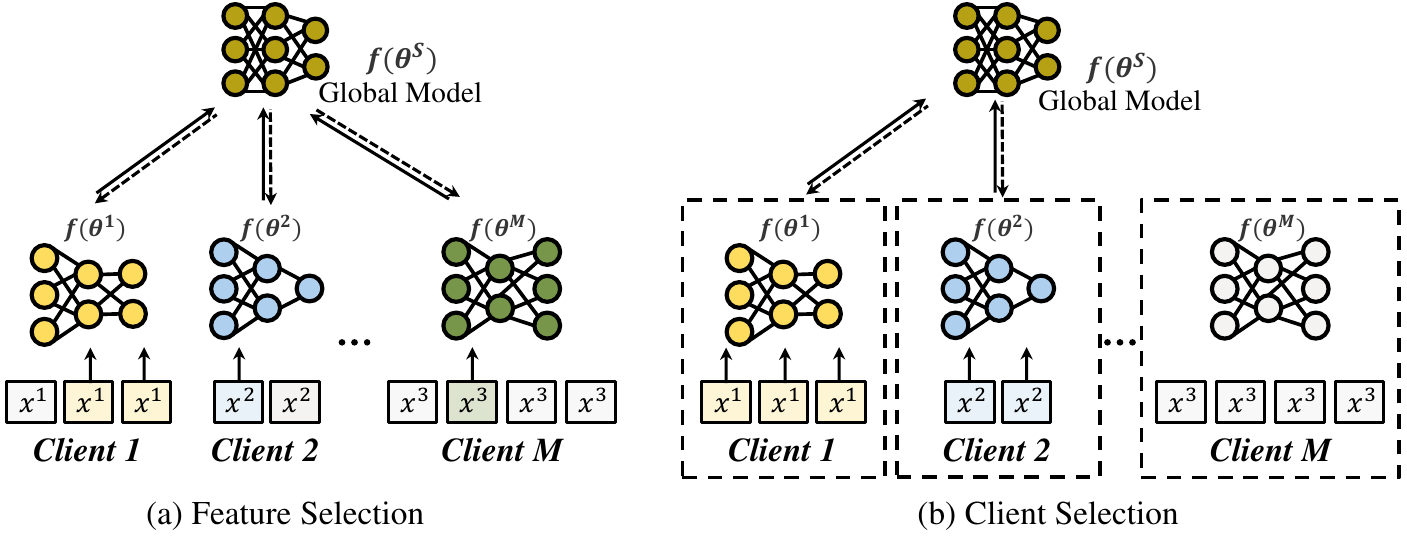}
    \caption{\textbf{The illustration of Feature \& Client Selection.} (a) Feature selection aims to identify crucial features for collaborative training across all clients. (b) Client selection involves choosing essential clients for collaboration.}
    \label{fig:Feature_Client_Selection}
\end{figure}

\subsection{Feature \& Client Selection}
In vertical federated learning, multiple participants contribute distinct features for common samples to perform prediction tasks. However, the high-dimensional raw features often contain irrelevant or noisy information that does not relate to the current prediction task. These redundant features, present in some clients, add unnecessary costs or introduce noise, hindering the effectiveness of vertical federated learning. Hence, selecting critical features and clients becomes crucial in vertical federated learning to ensure effectiveness. Traditional feature selection algorithms are designed for centralized data and may not directly adapt to vertical federated learning, where features are distributed across different clients. Additionally, while some works focus on client selection in federated learning, they are typically designed for shared feature spaces with different samples, rendering them unsuitable for vertical federated learning. To enhance effectiveness, recent research has proposed a variety of novel feature and client selection methods tailored specifically for vertical federated learning. These methods can be categorized based on the different selection criteria, which are detailed as follows:

\textbf{Feature Selection.} 
One strategy to enhance effectiveness is to select critical features for collaboration. Li \textit{et al.} \cite{li2023fedsdg} propose a dual gate-based feature selection method, efficiently approximating the probability of feature selection. They also introduce a local perturbation method for enhanced privacy. LESS-VFL \cite{castiglia2023less} explores and formalizes the feature selection problem for VFL, and achieves communication-effective feature selection using the group lasso algorithm. Feng \textit{et al.} \cite{feng2022vertical} propose feature selection without sample alignment through local weight regularizations. Zhang \textit{et al.} \cite{zhang2022secure} introduce a privacy-preserving feature selection method for electronic health, utilizing Gini impurity to measure feature importance and proposing customized protocols for different eHealth scenarios. Fu \textit{et al.} \cite{fu2023feast} propose FEAST, leveraging conditional mutual information to select informative features with low redundancy. PSO-EVFFS \cite{zhang2023embedded} integrates evolutionary feature selection into the SecureBoost framework, optimizing both hyperparameters of the XGBoost model and feature subsets to improve accuracy while ensuring data privacy. However, generalization to other backbones remains unexplored.

\textbf{Client Selection.} 
In addition to feature-level selection, several studies focus on client selection for effectiveness. FedSV \cite{wang2019measure} proposes to measure client contributions in VFL using Shapley Values. It calculates the Shapley value for each client during training and aggregates these values for the final contribution assessment. However, FedSV fails to achieve fairness as clients not selected in certain iterations are not considered. To address fairness concerns in client contribution measurement, Fan \textit{et al.} \cite{fan2022fair} propose VerFedVS, where client contributions are calculated across multiple time stamps, and embeddings from different timestamps are utilized to ensure fairness. Moreover, VerFedVS is communication-efficient and applicable in both synchronous and asynchronous settings. For efficient and secure client selection, Jiang \textit{et al.} \cite{jiang2022vf} propose VF-MINE, which utilizes mutual information (MI) to select a subset of clients that preserves maximum MI. Additionally, VF-MINE ensures privacy protection through homomorphic encryption and optimizes communication costs between participants.

\section{Security}
Security is a paramount concern in vertical federated learning, encompassing two critical aspects. Firstly, the threat of privacy leakage looms large, particularly during sample alignment and communication phases. Secondly, VFL faces the challenge of malicious attacks performed by evil parties. The attacks aim to access the private information of raw data or are designed to disrupt the collaborative learning process. As VFL involves multiple parties collaborating without sharing raw data, addressing these security concerns is essential not only for protecting the privacy of participants but also for constructing trust and reliability of the collaboration learning. In the following section, we illustrate the security issues in VFL into two aspects: Privacy Leakage and Malicious Attacks.

\subsection{Privacy Leakage}
Privacy leakage poses a significant challenge in vertical federated learning, manifesting across various stages of the collaborative process \cite{xu2022privacy,yu2024survey,jiang2022comprehensive}. The first concern is the information leakage during sample alignment, where shared samples between different clients become susceptible to unintentional exposure. Besides, the transportation of embeddings and gradients between clients introduces further vulnerabilities. Leakage in embedding transportation entails the risk of revealing sensitive information encoded in the intermediate representations, while leakage in gradient transportation raises the potential for malicious entities to infer private information during the collaborative model training. Addressing these concerns becomes imperative to ensure the security and confidentiality of the VFL systems. 

\textbf{Leakage in Sample Alignment.} 
Sample Alignment is the critical step for VFL, which aims to discover the common samples shared by the clients. However, privacy leakage during sample alignment presents a significant concern, arising from the necessity to compare a common identity or perform calculations with raw data. Sharing identities and raw data inadvertently introduces vulnerabilities that could compromise individual privacy of raw data. The fundamental challenge lies in achieving sample alignment without revealing private information, necessitating robust mechanisms and protocols to secure this process. To achieve privacy-preserving sample alignment, Hardy \textit{et al.} \cite{hardy2017private} propose the Privacy-Preserving Entity Resolution, where the raw data in clients is encrypted into a cryptographic long-term key (CLK). Subsequently, these CLKs are transmitted to the coordinator for matching entities, and the alignment information is then returned to the clients. It ensures the identification of common entities between different data providers without exposing private data. Lu \textit{et al.} \cite{lu2020multi} propose a multi-party private set intersection protocol that can handle cases where some parties drop out during the protocol execution. It relies on lightweight cryptographic primitives for efficient operation and can handle dropout scenarios without significant performance impacts. FLORIST \cite{sun2021vertical} utilizes Private Set Union (PSU) \cite{kissner2005privacy, davidson2017efficient, jia2022shuffle} to align data without revealing intersection membership. Besides, it employs synthetic data generation for samples outside the intersection to maintain privacy and model utility. However, the PSU protocol may increase the training sample size due to the union set, leading to higher training costs.

\textbf{Leakage in Embeddings Transportation.} 
In VFL, the intermediate embeddings are transmitted to the active client for training the global model. Privacy leakage in embedding transportation poses a critical concern in vertical federated learning, stemming from intermediate feature embeddings from each client inherently containing privacy-sensitive information derived from raw data. This issue becomes especially pronounced when considering potential malicious attackers or participants in collaboration. To address this challenge, Hardy \textit{et al.} \cite{hardy2017private} propose to leverage homomorphic encryption for intermediate embeddings, to ensure data privacy without leaking information from raw data. HDP-VFL \cite{wang2020hybrid} enables joint learning of a generalized linear model (GLM) from vertically partitioned data with minimal cost in terms of training time and accuracy. It leverages differential privacy (DP) to protect intermediate results (IR) during VFL, avoiding the need for Homomorphic Encryption (HE) or Secure Multi-Party Computation (MPC). However, under limited privacy budgets, the accuracy of the joint model may be constrained. Falcon \cite{wu2023falcon} supports VFL training for various machine learning models with strong privacy protection. It utilizes a hybrid strategy of threshold partially homomorphic encryption (PHE) and additive secret sharing scheme (SSS). Nevertheless, the requirements for threshold decryption in PHE and conversion to SSS can be computationally demanding.

\textbf{Leakage in Gradient Transportation.} During collaboration learning, the updated gradients are sent back to each client. Privacy leakage in gradient transportation represents a pivotal security challenge, wherein the gradients encompass sensitive information related to both the labels and the local models. These components, crucial for the collaborative learning process, are inherently private to each client. The transmission of gradients during collaboration introduces a potential avenue for privacy leakage, particularly when considering malicious attackers seeking to exploit the leakages. To address the privacy leakage with gradients, recent works utilize encryption or differential privacy paradigm to the VFL algorithm. FedV \cite{xu2021fedv} enables collaborative training of models without peer-to-peer communication among parties. It uses functional encryption to secure gradient computation for models like linear regression, logistic regression, and SVMs. However, the use of functional encryption may introduce computational complexity. CRDP-FL \cite{zhao2022vertically} integrates differential privacy into a VFL framework to protect privacy while maintaining utility. It ensures strong privacy guarantees by injecting differential privacy noise. AdaVFL \cite{errounda2023adaptive} adapts privacy protection based on feature contributions and model convergence. It uses zero-Concentrated Differential Privacy (zCDP) for privacy accounting, aiming to balance privacy protection and utility. Nonetheless, the protocol assumes honest-but-curious adversaries and secure communication channels, which may not be valid in all real-world scenarios. Besides, the adaptive budgeting scheme relies on the model convergence dynamics, which may vary with different datasets and tasks.

\begin{figure}
    \centering
    \includegraphics[width=0.85\linewidth]{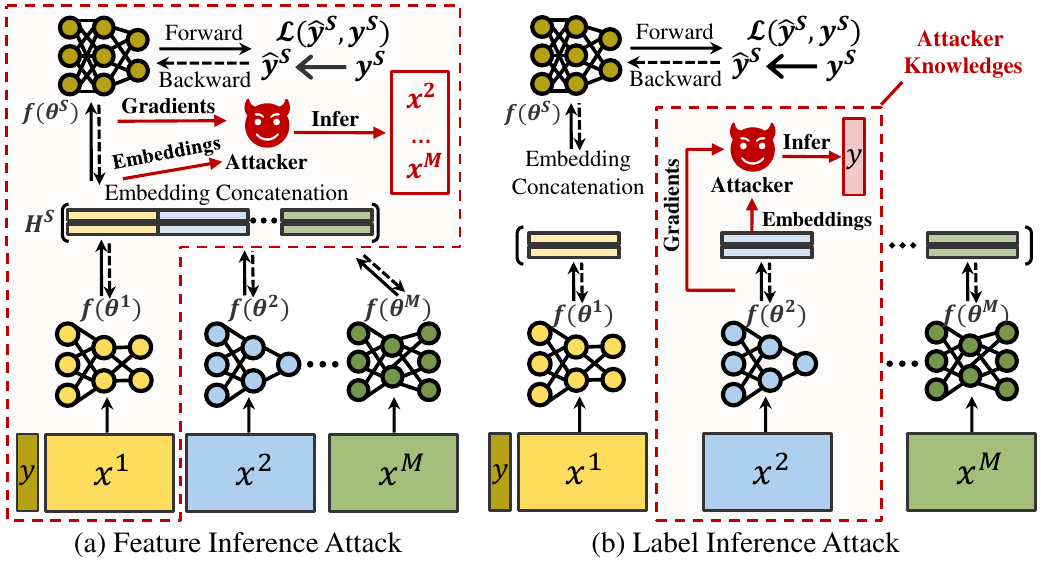}
    \caption{\textbf{Inference Attack.} (a) The attacker can infer the raw features using gradients and intermediate embeddings. (b) The attacker can infer the task labels using the trained local model and the gradients.}
    \label{fig:Inference_Attack}
\end{figure}

\subsection{Malicious Attack}
The malicious attack poses a significant threat to the security of VFL systems. It manifests in two primary forms: Inference Attack and Destructive Attack. In the former, malicious attackers attempt to intrude into the private domains of participants by exploiting information communicated during the collaboration. The goal is to infer raw data, directly threatening the confidentiality of participant-specific information. Besides, Destructive Attack poses a risk to the function and robustness of the learned model. Attackers attempt to disrupt the effectiveness and reliability of the learned model. In the following sections, we will examine these two aspects of malicious attacks, highlight their potential consequences, and critically, explore the corresponding defense strategies designed to enhance the security of VFL against such threats.

\subsubsection{Inference Attack} 
The Inference Attack poses a significant challenge to participant privacy in VFL, where attackers aim to recover private data by exploiting the information exchanged during collaboration. Through the analysis of exchanged information, attackers can reconstruct raw data from participants, representing a significant threat to system security. According to the obtained data, Inference Attack can be categorized into two types: Feature Inference Attack, where the goal is to uncover features using gradients or prediction outputs; Label Inference Attack, which aims to extract label information from the active client using gradient or model parameter information.

\textbf{Feature Inference Attack:} Luo \textit{et al.} \cite{luo2021feature} develop specific attacks that work with both single and multiple prediction outputs. They involve precise and generative inference and do not depend on any prior knowledge about the target data distribution, adaptable to various complex VFL models. However, the effectiveness of these attacks depends on the correlation between the features of the adversary and those of the passive parties. CAFE \cite{jin2021cafe} is designed to recover data in batches from shared aggregated gradients. It leverages data index and internal representation alignments to efficiently conduct large-batch data leakage attacks. Nevertheless, the effectiveness of CAFE is influenced by the model architecture and weight distribution initialization. Rassouli \textit{et al.} \cite{rassouli2022privacy} propose several inference attack techniques to reconstruct sensitive features from the passive party in VFL. They utilize the Chebyshev center concept for these attacks and provide theoretical performance guarantees, but computing the Chebyshev center is complex and computationally intensive. Ye \textit{et al.} \cite{ye2022feature} introduces an attack that attempts to reconstruct binary features from passive parties in VFL. It proves that while reconstructing general features is NP-hard, binary feature reconstruction is feasible and presents a search-based attack algorithm. However, this attack primarily targets binary features and may not be directly applicable to other types of data or non-binary features. Yang \textit{et al.} \cite{yang2023practical} propose to utilize an inference model to minimize the distance between predictions from inferred and target features. It employs zeroth-order gradient estimation to train the inference model due to the lack of direct access to the global model and other local model information. However, the effectiveness of the attack is influenced by the correlation between known features and target features. Higher correlations lead to better attack performance.

\textbf{Label Inference Attack:} Liu \textit{et al.} \cite{liu2021batch} introduce a gradient inversion model that can reconstruct private labels with high accuracy by exploiting batch-averaged local gradients. They also present a gradient-replacement attack that facilitates label replacement in black-box VFL settings without altering the existing VFL protocols. However, the success of this attack depends on the batch size being smaller than the dimension of the embedded features in the final fully connected layer. Li \textit{et al.} \cite{li2021label} propose a norm-based scoring function, which exploits the observation that the norm of the gradient vector can be different for positive and negative examples, potentially revealing label information. The scoring function uses the gradient norm as a predictor of the unseen label. Nonetheless, it is tailored to a specific threat model within two-party split learning, which may not generalize to other models or settings. Fu \textit{et al.} \cite{fu2022label} propose passive attacks using model completion, active attacks with a malicious local optimizer, and direct attacks by utilizing the gradient signs. However, the direct label inference attack is limited to training examples and requires additional steps for inference on new samples. Sun \textit{et al.} \cite{sun2022label} introduce a label inference attack method that exploits the correlation between intermediate embeddings and private labels to steal sensitive label information. However, the label party may need to increase computational resources to re-learn the correlation between embeddings and labels to maintain model utility. Exploit \cite{kariyappa2023exploit} frames the attack as a supervised learning task using gradient information obtained during split learning. Nevertheless, the efficacy of Exploit varies based on the chosen splitting layer. Further exploration could involve evaluating the attack performance across different splitting layers and examining its robustness to various defense mechanisms. Qiu \textit{et al.} \cite{qiu2022your} propose a numerical approximation method, which is designed to approximate encrypted representations in VFL, enabling the inference of private label-related relations. However, the effectiveness of the attack is contingent on access to certain knowledge, such as prediction results and global model parameters.

\begin{figure}
    \centering
    \includegraphics[width=1\linewidth]{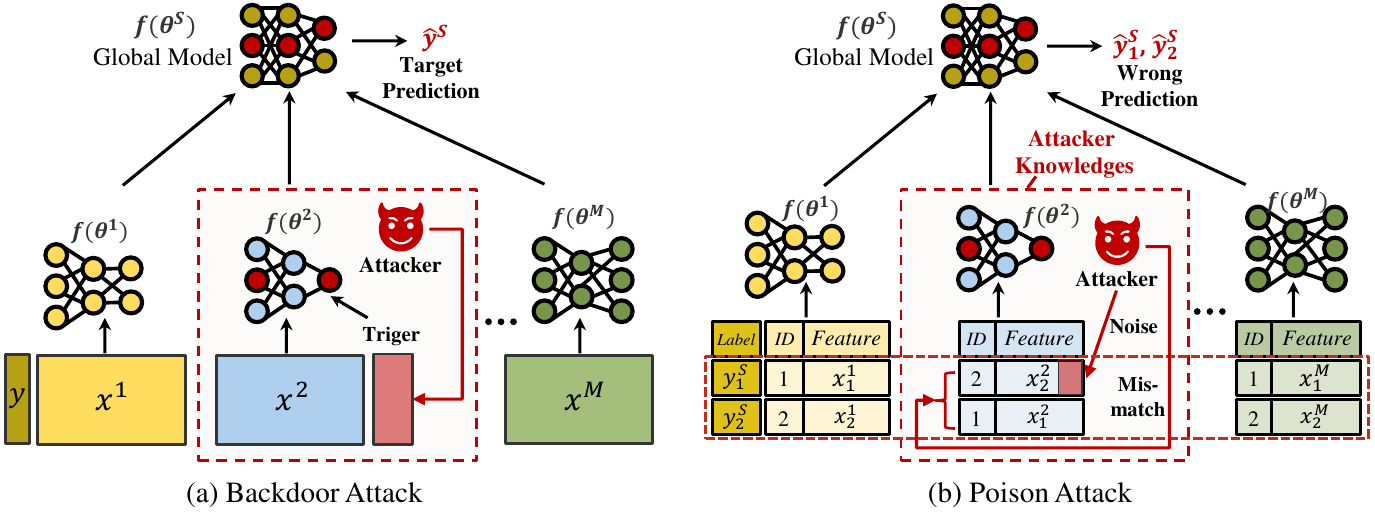}
    \caption{\textbf{Destructive Attack.} (a) The attacker can add a backdoor trigger to the data for making the target prediction. (b) The attacker can poison the raw data by adding noise or mismatching the sample IDs.}
    \label{fig:Destructive_Attack}
\end{figure}

\subsubsection{Destructive Attack}
In vertical federated learning, the destructive attack manifests when attackers attempt to compromise the prediction of the trained model. Attackers may employ various tactics, including injecting targeted modifications, introducing malicious data, manipulating model parameters, or disrupting the learning process. The Destructive Attack can be categorized into two types: \textit{Backdoor Attack} and \textit{Poison Attack}. The former involves surreptitiously injecting subtle modifications into the data, thereby causing the model to produce incorrect predictions for specific targets. Besides, the Poison Attack aims to corrupt the model or the underlying data directly, reducing the accuracy of model predictions and undermining the trustworthiness of the collaborative learning framework.

\textbf{Backdoor Attack:} Liu \textit{et al} \cite{liu2020backdoor} propose to record the intermediate gradient of a clean sample with the targeted label and use this recorded gradient for poisoned samples. The goal is to make the model assign an attacker-chosen label to input data with a specific pattern or trigger. However, the modification of gradients will influence the performance on clean data. UAB \cite{chen2024universal} utilizes bi-level optimization to alternately optimize universal backdoor trigger generation and model parameters during VFL training, without requiring additional data or labels from other participants. However, the attack performance will be affected if the dataset is imbalanced. LR-BA \cite{gu2023lr} involves an adversarial participant in VFL who fine-tunes their local model to produce specific latent representations for backdoor instances, thereby manipulating the federated model to predict an attacker-specified label. It can be executed even without access to labels, using only local latent representations, but its performance may vary with the dataset and number of classes. He \textit{et al.} \cite{he2023backdoor} propose to utilize a malicious client injecting a stealthy backdoor into the global model during training. This is achieved by replacing the local embeddings of a small number of target-class samples with a carefully constructed trigger vector, without modifying any labels. However, the attack method is non-trivial and requires careful construction of the trigger vector. VILLAIN \cite{bai2023villain} introduces a novel label inference algorithm to insert a backdoor into the global model. It intensifies the backdoor attack power by designing a stealthy additive trigger and introducing backdoor augmentation strategies to impose a larger influence on the global model. Nevertheless, its effectiveness relies on the sophisticated label inference algorithm, which introduces additional complexity. BadVFL \cite{xuan2023practical} introduces a Source Data Detection (SDD) module to trace data categories based on gradients and a Source Data Perturbation (SDP) scheme to enhance the decision dependency between the trigger and attack target. Nonetheless, the effectiveness of the attack may vary with the complexity of the global model structure, affecting gradient-based calculations. Graph-Fraudster \cite{chen2022graph} leverages noise-added global node embeddings and gradients of pairwise nodes to confuse the global model. However, the success of the attack relies on the leakage of global node embeddings.

\textbf{Poison Attack:} VFedAD \cite{lai2023vfedad} proposes three data-level poison attack methods: Random Failure, which introduces noise to raw data; Random Mismatch, where IDs in some clients are shuffled to disrupt collaborative learning performance; and Targeted Tampering, where the attacker replaces some samples with target features. However, the attacks are easy to detect due to the conspicuous modification of raw data. Qiu \textit{et al.} \cite{qiu2024hijack} introduce two attacks, the replay attack and the generation attack, demonstrating the vulnerability of VFL systems to the Byzantine Generals Problem. It also explores a poisoning phase to enhance these attacks when the adversary has limited features. Yao \textit{et al.} \cite{duanyi2023constructing} propose an attack that disrupts the VFL inference process by adaptively corrupting a subset of clients. It formulates finding optimal attack strategies as an online optimization problem, involving adversarial example generation and corruption pattern selection. However, the ability to adjust corruption patterns relies on the effectiveness of previous attacks, which may not always lead to the discovery of the optimal pattern. P-GAN \cite{chen2024gan} utilizes semi-supervised learning to create a surrogate target model, then employs a GAN-based method to generate adversarial perturbations that degrade the model performance. However, the effectiveness is influenced by various factors like the number of attacker features and the known labels.

\subsubsection{Defense}
Defending against malicious attacks is crucial for ensuring the security of VFL. Two primary defense strategies involve mitigating inference attacks and thwarting destructive attacks. Mitigating the risks posed by inference attacks, which seek to infer sensitive data from transmitted information, requires the implementation of sophisticated encryption and privacy-preserving protocols. Besides, countering destructive attacks demands measures to detect and neutralize malicious attempts aimed at undermining the integrity and functionality of learned models.

\textbf{Defense Against Inference Attacks.}
For \textit{defending against feature inference attacks,} Sun \textit{et al.} \cite{sun2021defending} propose an adversarial training framework with three modules: adversarial reconstruction, noise regularization, and distance correlation minimization, which work together to make the collaboration model robust against attacks that attempt to reconstruct sensitive input data from shared gradients. Nevertheless, the adversarial nature of the training process adds complexity and may require careful tuning to achieve the desired balance between privacy and performance. VFLDefender \cite{zhu2024vulnerabilities} employs gradient obfuscation to reduce the correlation between model updates and training data, effectively preventing reconstruction attacks and preserving data privacy. However, the defense mechanism impacts model utility and the balance between privacy and performance needs further exploration. Chang \textit{et al.} \cite{chang2024gradient} propose to selectively transmit only a portion of the gradient components, reducing the risk of data leakage while maintaining model training accuracy. Besides, they use cosine similarity to measure the angular difference between client-updated gradients and server-sent gradients. Clients can then disguise their uploaded gradient to protect privacy without compromising the global model accuracy. However, it requires additional processing of gradient components, which can be computationally intensive. Mao \textit{et al.} \cite{mao2022secure} implement a fine-grained privacy budget allocation scheme to efficiently perturb information exchange between the passive and the active client, effectively protecting from data reconstruction attacks. FedPass \cite{gu2023fedpass} introduces an adaptive obfuscation mechanism that adjusts during the learning process to protect features and labels simultaneously. It embeds private passports in both passive and active party models, making it exponentially hard for attackers to infer features and nearly impossible to infer private labels. However, the randomness in passport generation is crucial for data privacy, but the process and its security implications require careful consideration. 

For \textit{defending against label inference attack,} Li \textit{et al.} \cite{li2021label} proposes minimizing label leakage in split learning by optimizing noise perturbation structures. It strategically adds random noise to gradients and prevents adversaries from recovering private labels. However, it assumes a gaussian distribution for unperturbed gradients and can be computationally intensive with multiple optimizations. DCAE \cite{zou2022defense} proposes to defend against label inference and replacement attacks by disguising true labels using autoencoder and entropy regularization with prior label knowledge. However, the implementation of DCAE may add complexity to the VFL system. Sun \textit{et al.} \cite{sun2022label} propose an additional optimization goal at the active client to minimize the distance correlation, making it difficult for an adversary to infer private labels from the shared intermediate embedding. However, the method is primarily evaluated in a binary classification setting, and its effectiveness in other scenarios or with different data distributions is not extensively discussed. Besides, the active client may need additional computational costs to re-learn the correlation between embeddings and labels. FLSG \cite{fan2023flsg} defends against passive label inference attacks by generating gradients similar to the original ones using a Gaussian distribution. It measures the cosine distance between the generated and original gradients and selects the most similar ones to replace the original gradients during back-propagation. Despite being lower than other methods, implementing FLSG still increases the training time compared to the original VFL framework. Wang \textit{et al.} \cite{wang2023beyond} introduce a shadow model to share gradients during training, disrupting the correlation between gradients and training data, which hinders attackers from inferring labels. However, the introduction of a new local model and the need for secret sharing during training add complexity and potential cost to the process. ProjPert \cite{fu2024proj} addresses the issue of label leakage during training by formulating an optimization problem that minimizes the impact on model quality while satisfying a pre-set privacy guarantee. However, the heuristic variant may not always provide the optimal solution but is designed to be close to it. HashVFL \cite{qiu2024hashvfl} addresses the challenges of learnability, bit balance, and consistency in VFL. It uses a sign function for binarization, batch normalization for bit balance, and predefined binary codes for consistency. However, the integration of hashing can cause vanishing gradients during model training. Zheng \textit{et al.} \cite{zheng2022making} propose the use of potential energy loss (PELoss) to make the output distribution of the local model more complex, protecting against label leakage. It pushes outputs of the same class toward the decision boundary, making it difficult for an attacker to fine-tune the local model with a small number of leaked labeled samples. However, the method assumes the attacker uses supervised learning to fine-tune the local model and does not consider attacks based on unsupervised learning approaches. TPSL \cite{yang2022differentially} adds noise to gradients and model updates during training to ensure differential privacy, focusing on protecting sensitive label information. Takahashi \textit{et al.} \cite{takahashi2023eliminating} propose defense mechanisms to mitigate label leakage in tree-based VFL, by utilizing label differential privacy with post-processing and mutual information regularization. Nevertheless, it requires additional training and communications.

\textbf{Defend Against Destructive Attacks.}
He \textit{et al.} \cite{he2023backdoor} propose strategies from three aspects: Statistical Filtering, which filters out abnormal local embeddings that deviate from natural outputs, but may not detect stealthy attacks with well-crafted trigger vectors; Reverse Engineering Detection, identifies backdoored models by comparing reversed trigger vectors for different classes, but with low detection rates, indicating difficulty in identifying malicious models; Confusional Autoencoder, which maps original labels to fake labels to minimize classification probability differences, but does not significantly hinder the proposed attack, as it does not rely on gradients or label inference. Chen \textit{et al.} \cite{chen2024gan} focus on a server-side anomaly detection algorithm based on a deep auto-encoder (DAE) to defense against data poisoning attacks. The DAE method is used to identify outliers in embedding vectors with reconstruction errors and filter out anomalous data, ensuring that only clean training data is used on the server. Nonetheless, the detection might struggle with high proportions of poisoned data;  VFedAD \cite{lai2023vfedad} uses information theory to detect anomalies in vertical federated learning. It employs contrastive learning and cross-client prediction tasks to learn data representations that help identify poisoned samples. Nonetheless, the success of anomaly detection hinges on the quality of learned data representations. Qiu \textit{et al.} \cite{qiu2024hijack} utilize two kinds of strategies to defend the destructive attack: Normalization, which is effective against black-box attacks like ZOO by transforming perturbed input, preventing gradient approximation. However, the impact is limited when the attacker has a significant feature ratio, as the threat persists; Dropout reduces model memorization of specific patterns, which can mitigate attacks, but high dropout probability can drastically reduce main task performance, making it impractical for collaborative learning. RVFR \cite{liu2021rvfr} operates by training individual feature extractors in quarantine, followed by a robust feature subspace recovery process, and feature purification based on the assumption that only a small fraction of agents are malicious. The framework is theoretically grounded, providing guarantees for the recovery of uncorrupted features and the robustness of the model against a range of attacks. However, the effectiveness of RVFR hinges on specific assumptions such as the existence of a low-rank feature subspace and the predominance of well-intentioned agents.

\section{Applicability}
Vertical Federated Learning (VFL) has emerged as a pivotal technology in a multitude of application domains \cite{yang2019federated, liu2024vertical, zhang2021vertical, wei2023fedads, yan2024cross, huang2023vertical, li2020review, liu2021fate}. Despite its advances, the practical deployment of VFL systems frequently encounters several obstacles that can significantly hinder their applications. Recent research identifies challenges from three main aspects: (1) \textit{Limited Data.} Obtaining sufficient high-quality data for model generalization is crucial, but VFL faces limitations in data availability. This includes: (i) Limited Aligned Samples: VFL relies on all clients possessing parts of features from the same samples. However, the number of common samples is often limited, hampering model generalization. (ii) Limited Labels: In many cases, collecting high-quality labels is challenging, which reduces collaboration performance. (2) \textit{Large Communication Burden.} The communication cost in VFL is significant, as intermediate features and updated gradients are frequently exchanged. However, device resources are typically limited. Addressing this challenge is crucial for deploying VFL systems. (3) \textit{Client Asynchrony.} In practice, the device and computing power differences cause the calculation of intermediate results to be delayed in some clients, which affects the updation of collaborative learning. It shows the asynchrony in vertical federated learning and becomes a critical issue in achieving the applicability of the VFL system.

\begin{figure}
    \centering
    \includegraphics[width=0.92\linewidth]{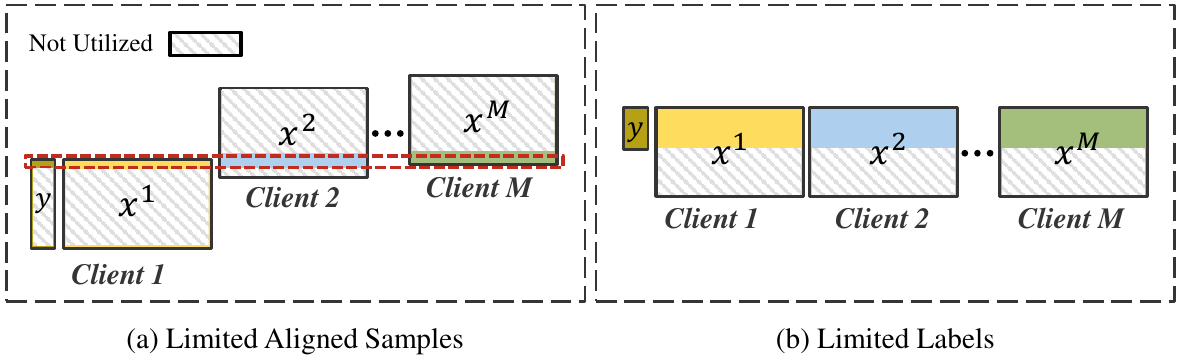}
    \caption{\textbf{Limited Data.} (a) Aligned samples are often limited, especially as the number of clients increases. (b) The task labels in the active client are limited, leaving unlabeled samples underutilized.}
    \label{fig:Limited_Data}
\end{figure}

\subsection{Limited Data}
The amount of high-quality data is critical to improve the generalization of the trained models. However, a large amount of sufficient data is usually unavailable in VFL, which impedes the applicability of VFL. The challenges of limited data can be divided into two aspects:

\textbf{Limited Aligned Samples.}
The performance of Vertical Federated Learning (VFL) heavily relies on the availability of common samples shared across different clients for collaborative training. However, in practice, the number of aligned samples is often restricted, particularly as the number of clients increases. This limitation in the size of aligned samples impedes collaboration performance and diminishes the applicability of VFL. Recent efforts have extensively explored solutions to enhance the applicability of VFL in the face of limited aligned samples. They strive to address this challenge by maximizing the utilization of unaligned samples through feature generation and pseudo-label estimation. For instance, FedCVT \cite{kang2022fedcvt} proposes to complete missing features for unaligned samples and assign pseudo-labels, effectively expanding the sample size for collaborative training. Sun \textit{et al.} \cite{sun2023communication} introduced a semi-supervised paradigm in which each client updates with estimated labels to improve sample utilization. However, these methods face various limitations. Firstly, both label estimation and feature generation are prone to errors, introducing noisy information. Moreover, these methods exclusively leverage samples with high-confidence pseudo-labels for learning, leaving a significant number of samples underutilized. Furthermore, accurately generating pseudo-labels and features becomes even more challenging in scenarios with limited aligned samples due to the reliance on prior information about aligned samples. FedMC \cite{yang2022multi} expands the training data by matching non-overlapping samples based on similarity, thus improving the effectiveness of the jointly trained model. However, it aims to pair the unaligned samples upper the desired similarity for collaboration training, which cannot fully utilize the samples. Given these limitations, future research could explore more efficient frameworks to overcome the challenges posed by limited aligned samples. 

\textbf{Limited Labels.}
Limited labels pose a significant challenge to the applicability of VFL. In vertical federated learning, multiple clients collaborate to build a shared model for the collaboration task, with the labels from the active client. However, the availability of labeled data for training is often constrained due to various reasons, including privacy concerns and the difficulty of providing large amounts of high-quality labeled data. The scarcity of labeled data hampers collaborative training, impacting the performance of the global model. To improve applicability in situations with scarce labels in VFL, recent works propose a series of unsupervised and self-supervised methods to leverage unlabeled data. For instance, Cha \textit{et al.} \cite{cha2021implementing} present a simple and robust VFL paradigm based on auto-encoders, which does not require domain knowledge or labels to train the autoencoder models. However, the proposed method transforms data into a high-dimensional latent space, introducing redundant information and potential overfitting on short data sequences. FedOnce \cite{wu2022practical} proposes to enhance the intermediate features by unsupervised learning in each client without labels. However, labels for the collaboration task are still needed for training; Besides, additional computational costs are introduced with local updation. FedHSSL \cite{he2024hybrid} presents a pre-train method for VFL. The basic idea is to leverage cross-party views, local views, and invariant features of samples to improve the performance of VFL collaboration with limited labels. However, it relies on the assumption that participating parties have similar data distributions, and additional communication costs are introduced, which may affect scalability and efficiency. SS-VFL \cite{castiglia2022self} leverages unlabeled data to train representation networks and labeled data for a downstream prediction network, aiming to achieve higher accuracy with reduced communication costs. However, the effectiveness may vary depending on the similarity between local and centralized class probability distributions. Additionally, there is a potential for model bias if the datasets used for training are biased.

\begin{figure}
    \centering
    \includegraphics[width=0.88\linewidth]{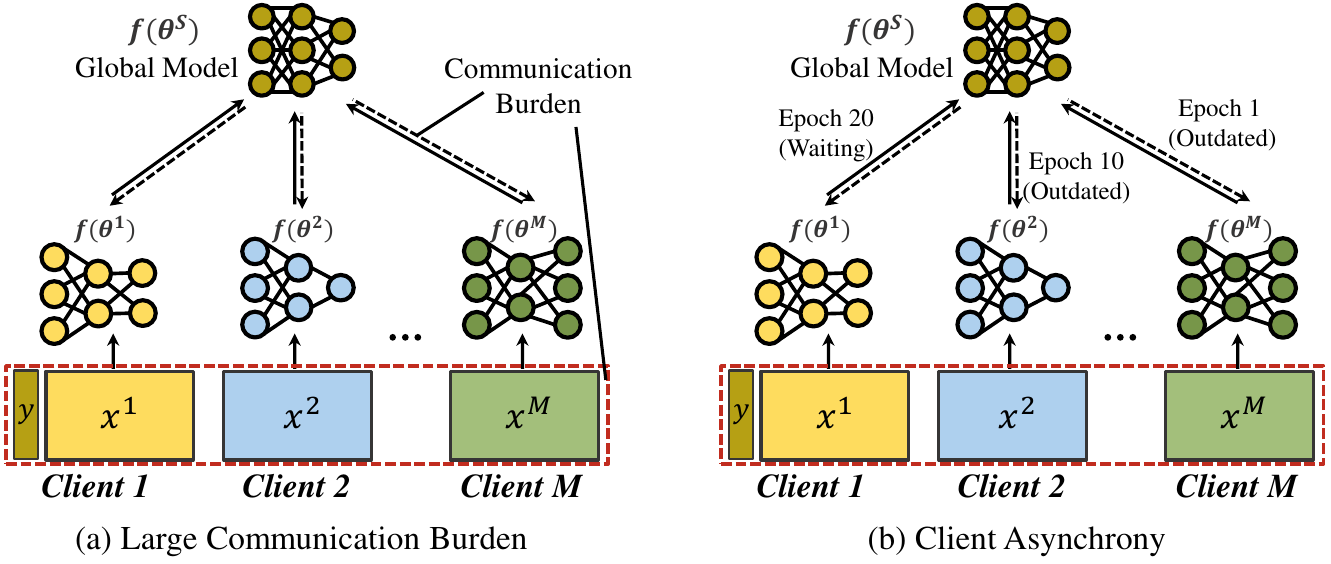}
    \caption{\textbf{Large Communication Burden and Client Asynchrony.} (a) The sample alignment process and multiple communication rounds introduce significant communication burdens. (b) Differences in computational power and bandwidth cause the asynchronous updates between clients.}
    \label{fig:Communication_Asynchrony}
\end{figure}

\subsection{Large Communication Burden}
The challenge of a large communication burden in Vertical Federated Learning arises from the need to exchange substantial amounts of information across different clients during the collaboration process. In VFL, each client retains its local dataset and models, then the global model is built with intermediate features and collaboration labels. During this process, the communication of intermediate information such as features, gradients, and sample alignment indices between clients can be resource-intensive, especially when dealing with large-scale datasets or models. The sheer volume of data that needs to be transmitted over networks introduces latency and consumes bandwidth. This challenge is particularly pronounced in scenarios where the participating clients are geographically dispersed or have limited communication resources, which hinders the application of VFL. To mitigate the large communication burden and improve the applicability of VFL, many innovative strategies have been proposed in recent works. AsySQN \cite{zhang2021asysqn} proposes to use asynchronous stochastic quasi-Newton methods to train VFL models. It leverages approximate second-order information to reduce the number of communication rounds and improve the convergence speed. However, it requires the collaboration labels to be held by all parties, which may not be realistic in some scenarios. Khan \textit{et al.} \cite{khan2022communication} propose utilizing feature compress methods to compress the local data of each party into latent representations and reduce the communication rounds into one, thus reducing the total communication overhead. However, it depends on the choice of the feature extraction technique, which may affect the quality of the latent representations and the final model. FedBCD \cite{liu2022fedbcd} utilizes a Federated Stochastic Block Coordinate Descent algorithm, enabling parties to perform multiple local updates before each communication, reducing communication overhead. However, the number of local-update rounds needs careful selection to avoid potential divergence. Besides, the performance can be sensitive to the choice of stepsize and batch size. C-VFL \cite{castiglia2022compressed} leverages several local updations with compressed embedding sharing, which reduces the communication cost by over 90\% compared to VFL without compression, without performance degradation. Nevertheless, it requires bounded embedding gradients and bounded Hessian of the objective function, which may not hold for some models or datasets. CELU-VFL \cite{fu2022towards} utilizes cached stale statistics to estimate model gradients for local updation, reducing the need for frequent cross-party communication. However, it involves approximation with stale statistics, which may introduce errors in gradient estimation. SparseVFL \cite{inoue2023sparsevfl} reduces the data size exchanged between the server and clients by exploiting the sparsity of embeddings and gradients. However, the sparsity may vary depending on the data distribution, neural network architecture, and hyperparameters, making it less effective for dense embeddings/gradients or high-dimensional embeddings/gradients. VFL-CZOFO \cite{wang2024unified} applies a cascaded hybrid optimization method that uses a zeroth-order (ZO) gradient on the most critical output layer of the clients, and a first-order (FO) gradient on the other parts of the model. It greatly alleviates the slow convergence problem of ZO-based VFL and significantly reduces communication costs compared with state-of-the-art communication-efficient VFL frameworks. However, it increases computational costs, requiring multiple forward propagations on its local model, and requires a trade-off between compression rate and test accuracy. Cheetah \cite{huang2022cheetah} introduces homomorphic encryption-based protocols for linear layers without expensive rotation operations and efficient primitives for non-linear functions, resulting in faster and more communication-efficient secure two-party neural network (2PC-NN) inference. However, the protocol involves multiple steps, including partitioning input shares, encoding to polynomials, homomorphic operations, and re-masking, which can be computationally intensive. FDSKL \cite{gu2020federated} uses random features to approximate kernel mapping functions and doubly stochastic gradients for solution updates, all computed without revealing data. The algorithm is shown to be faster and more communication-efficient than existing federated learning methods when dealing with kernels while maintaining similar generalization performance. However, the model coefficients are stored separately across different workers to maintain privacy, which could complicate the model management and updating process.

\subsection{Client Asynchrony}
Client asynchrony in VFL arises because participating clients operate on their local datasets and communicate intermediate results independently with different computational powers and bandwidth, potentially leading to temporal misalignment in the collaboration process. Unlike synchronous scenarios where all clients update their models simultaneously, asynchronous VFL involves clients uploading intermediate features and updating models at different times. This asynchrony introduces complexities in maintaining consistency and coordination among the models. Issues such as parameter staleness, where some models may use outdated information, can arise, impacting the convergence of the global model. Recent works have widely explored solutions to address this challenge. VAFL \cite{chen2020vafl} allows clients to run stochastic gradient algorithms without coordination. Clients can participate intermittently or strategically, but it may become complex when dealing with nonlinear local embedding functions. Li \textit{et al.} \cite{li2020efficient} propose an efficient method for client asynchrony, introducing gradient prediction using second-order Taylor expansion and double-end sparse compression to reduce training time and network traffic without sacrificing accuracy. However, it requires synchronous warm-up training at the beginning, and the gradient prediction method may introduce estimation errors, although they are mitigated by adaptive hyper-parameters. Gu \textit{et al.} \cite{gu2021privacy} introduce an asynchronous federated stochastic gradient descent (AFSGD-VP) algorithm and its variants for vertically partitioned data. It improves efficiency by keeping all computational resources busy, avoiding idle time waiting for synchronization. Although it addresses the straggler problem well initially, as the number of workers grows, the communication overheads can limit the scalability and speedup of the algorithms. VFB$^{2}$ \cite{zhang2021secure} integrates a backward updating mechanism and bilevel asynchronous parallel architecture, enabling asynchronous model updates across parties. However, VFB$^{2}$ assumes that one or partial parties hold labels, which may not be the case in all real-world applications. Shi \textit{et al.} \cite{shi2022practical} propose to leverage asynchronous training to reduce waiting time and use secret sharing instead of HE for privacy protection, eliminating the need for a trusted VFL coordinator. Nevertheless, the total computational complexity and communication cost can increase with the number of participants and the size of the data. vFedSec \cite{qiu2023vfedsec} employs a robust aggregation mechanism that can handle missing or temporarily disconnected clients. Instead of waiting for all clients to participate in each round, vFedSec allows partial model updates. Besides, vFedSec adjusts the learning rate dynamically based on the number of active clients. However, it might require significant computational resources, making it less practical for real-time applications. Sun \textit{et al.} \cite{sun2023robust} propose party-wise dropout to prevent the co-adaptation of feature extractors across parties, which is the root cause of performance drops. However, the method addresses the co-adaptation of feature extractors across parties, but it doesn’t eliminate it entirely. There’s still a chance that the model may rely on certain representation components from specific parties’ extractors. FedVS \cite{li2023fedvs} is designed to handle straggling clients during training without performance degradation. It uses secret sharing schemes for local data and models, ensuring that the aggregation of all client embeddings is reconstructed losslessly from non-straggling clients. However, the feature reconstruction may have limitations while the non-straggling clients are in limited numbers.

\section{Future Research Directions}
In this section, we discuss the potential research directions that remain unexplored and can promote advances in the future.

\subsection{Effectiveness/Applicability and Security Trade-off}
In the domain of vertical federated learning (VFL), there is a complex interplay between striving for effectiveness and applicability on one hand, and the imperative to enhance security on the other. This balance characterizes the Effectiveness/Applicability and Security Trade-off, where efforts to enhance the security of the system may impact its efficacy and usability. This trade-off issue has been widely discussed in horizontal federated learning (HFL) \cite{zhang2022no, li2024clients, zhang2023trading, girgis2021shuffled, chen2023privacy}. As VFL endeavors to harness disparate data sources while ensuring privacy, trade-offs inevitably arise. Tran \textit{et al.} \cite{tran2023privacy} offer a theoretical characterization of the relationship between privacy, convergence error, and communication cost. Kang \textit{et al.} \cite{kang2022framework} propose an evaluation strategy to measure the trade-off performance between privacy and utility, which offers references for future works. Understanding and navigating the complex trade-offs in VFL is essential for fostering the advancement and adoption of a secure, resilient, and effective collaborative system.

\subsection{Effectiveness Facilitates Security and Applicability}
In VFL, effectiveness serves as the cornerstone for enhancing both security and applicability. An effectively designed model not only ensures improved computational efficiency but also broadens the scope of applicability by catering to diverse and complex data scenarios. Besides, through feature \& client selection, it will allow for the prioritization of significant features or critical clients, further streamlining the learning process, and reducing computational overhead for enhanced applicability. Concurrently, the feature \& client selection process plays a pivotal role in identifying and mitigating potential vulnerabilities, enabling the early detection of attack vectors. By intertwining effectiveness with model design and feature \& client selection, a harmonious balance is achieved for enhanced applicability and security.

\subsection{Open Issues}
In Vertical Federated Learning (VFL), several open issues remain that are crucial for advancing the field yet have not been fully addressed. One major challenge is the absence of large-scale practical datasets. While current studies often evaluate their approaches by dividing data by features, there is a lack of standardized datasets that match practical VFL scenarios, resulting in biased evaluations. Additionally, the robustness of VFL systems against data noise has not been thoroughly investigated. Data noise can significantly impact the accuracy of models, yet strategies to mitigate this effect in VFL are underexplored. Another critical issue is the potential for unfair prediction bias towards certain features, which can lead to unequal treatment of different data inputs and raise ethical concerns. Furthermore, the exploration of VFL in handling various data variants, such as multi-modal data and graph data, is still in its infancy. These data types are increasingly common in practical applications, and their integration into VFL systems poses unique challenges and opportunities. Finally, with the development of foundation models, how vertical federated learning can facilitate their advancements is also a significant topic. Addressing these open issues is vital for the development of more robust, fair, and versatile VFL systems.

\subsubsection{Practical Datasets}
In VFL, different clients possess subsets of features on aligned samples, and cross-domain collaboration is conducted without leaking private data. A dataset collected from practical cross-domain collaborations would significantly facilitate VFL research. However, due to privacy concerns, a large-scale practical dataset for VFL is lacking. Existing works evaluate with data partitioned on standard datasets, such as dividing images or tabular data into segments and distributing them to clients. The VFLAIR \cite{zou2023vflair} benchmark evaluates 11 attacks and 8 defenses across 9 datasets, including image datasets \cite{lecun1998gradient, krizhevsky2009learning, chua2009nus, street1993nuclear} and tabular datasets \cite{adult2, dua2017uci, diabetes}, with data equally partitioned among each client. However, feature partitioning compromises data integrity and deviates from practical scenarios. FedSim \cite{wu2022coupled} proposes linking multiple datasets based on similarity measurements to simulate practical scenarios. However, the linkage method does not apply when each sample is uniquely identified. Vertibench \cite{wu2023vertibench} proposes to evaluate the distribution of data partitions based on party importance and correlation, and introduces a real-world image-to-image VFL dataset named Satellite, capturing satellite imagery of the same locations from 16 different visits. Despite its advancement, datasets where clients hold related but distinct data are lacking. The collection of practical datasets will be a critical focus for future research.

\subsubsection{Robustness and Generalization} 
\textit{Robustness} in Vertical Federated Learning (VFL) is crucial to ensure models can handle challenges such as data noise. Data noise, which encompasses inaccuracies or errors stemming from various sources, can lead to incorrect predictions by the model. A robust VFL system is designed to resist such disturbances, guaranteeing that the learned model remains reliable even with imperfect data. Besides, the \textit{Generalization} with unfair biases in prediction is equally important. It involves ensuring that the model does not deviate to certain features, which could result in biased predictions and compromise fairness across all data points. To enhance prediction fairness, FairVFL \cite{qi2022fairvfl} proposes to improve fairness in VFL models by learning unified and fair representations of data in a privacy-preserving manner. It employs adversarial learning to eliminate bias in data representations and utilizes contrastive adversarial learning to ensure user privacy. By removing bias from data representations, FairVFL promotes fairness across different user groups. Liu \textit{et al.} \cite{liu2021achieving} propose to model the fair learning task as a nonconvex-constrained optimization problem, aiming to discover a model that minimizes the loss function while adhering to fairness constraints. Formulating the problem in this way explicitly incorporates fairness considerations into the learning process. 

\subsubsection{VFL on Different Data Variants}
Extending vertical federated learning to different data variants such as multi-modal and graph data holds significant promise for practical applications, enabling more comprehensive information aggregation by leveraging diverse data sources. Multi-modal data amalgamates various types of information, including text, images, and audio, providing a holistic view of the data. Gong \textit{et al.} \cite{gong2023multi} develop a vertical federated learning framework to handle multi-modal data distribution challenges. Graph data, on the other hand, captures relationships and networks, providing deep insights into the interconnected nature of data points \cite{chen2020vertically, cheung2021fedsgc, ni2021vertical, chen2022graph, zhang2023glasu, mai2023vertical, wu2023privacy}. By integrating VFL with different data variants, we unlock new potentials for solving complex problems, such as understanding social networks and enhancing multimedia recommendations. It not only expands the applicability of VFL but also lays the groundwork for more sophisticated and inclusive data analysis techniques.

\subsubsection{VFL with Foundation Models}
Recently, foundation models \cite{bommasani2021opportunities, yuan2021florence, chang2024survey} have shown promising results and applications in various scenarios, such as GPT models \cite{radford2019language, brown2020language, achiam2023gpt}, LLaMA \cite{touvron2023llama}, and Sora \cite{videoworldsimulators2024}. While training foundation models requires a large amount of high-quality data, federated learning provides an alternative method to collaborate data from multiple clients without privacy leakage. Although several works have explored the combination of federated learning and foundation models \cite{zhuang2023foundation, fan2023fate, charles2024towards, yu2023federated, guo2023promptfl}, there is a lack of comprehensive discussion about vertical federated learning with foundation models. When deploying foundation models, there are privacy leakages concerning both data and model parameters. With VFL, multiple clients collaborate to train a prediction model, where only intermediate embeddings are transmitted to protect data privacy. Additionally, the entire model can be split among different clients, thereby eliminating the risk of leaking the complete model parameters. Moreover, the data can be enhanced in the feature space, which complements the data utilization in sample spaces. To explore the utilization of VFL with foundation models, Zheng \cite{zheng2023input} proposes using vertical federated learning as a privacy protection method for training foundation models and constructs a basic flow for training foundation models in VFL settings, addressing concerns about reconstruction attacks. However, training foundation models requires multiple training epochs and large data scales, which introduces significant communication and computation burdens. In the future, it is significant to explore the efficient deployment of VFL with foundation models.

\section{Conclusion}
In this survey, we provide a comprehensive and systematic review of recent progress in vertical federated learning. It is the first survey to summarize related works from the perspectives of effectiveness, security and applicability, presenting the most extensive and up-to-date literature. Initially, we provide a brief introduction to vertical federated learning, including the basic training protocols of VFL, equipping readers with essential background knowledge. Subsequently, we explore research topics from three perspectives: Effectiveness, focusing on developing effective learning paradigms for general settings in VFL; Applicability, which aims to explore solutions for various limited scenarios in VFL with potential applications; and Security, which ensures privacy and resistance to malicious attacks. The relevant works are summarized accordingly. Finally, we provide an outlook on valuable future research directions, which will facilitate advancements in vertical federated learning. In conclusion, vertical federated learning is attracting significant attention in related research fields and has made rapid progress recently. However, numerous research challenges remain unexplored.

\bibliographystyle{ACM-Reference-Format}
\bibliography{survey}

\newpage
\appendix
\onecolumn
\section{Appendix}
\begin{table*}[h]
\centering
\footnotesize
\caption{\textbf{Works on Model Design.}}
\label{tab:model_design}
\tabcolsep 2pt 
\begin{tabular}{|c|c|c|m{4.8cm}|m{4.8cm}|} 
  \cline{1-5}
  \multicolumn{2}{|c|}{Topics} & \makecell[c]{Ref.} & \makecell[c]{Key Contributions} & \makecell[c]{Limitations} \\
  \cline{1-5}
  \multicolumn{2}{|c|}{~}  & \makecell[c]{\cite{wu2020privacy, cheng2021secureboost, chen2021secureboost+, feng2019securegbm, fang2021large}} & Explore privacy-preserving strategies for IDs and feature embeddings, which intuitively achieve vertical federated learning, providing the foundation for effective collaboration. & They rely on specific designs for certain scenarios and lack a general solution to construct effective collaboration models. \\
  \cline{3-5}
  \multicolumn{2}{|c|}{~}  & \makecell[c]{\cite{tian2020federboost, li2022opboost}} & Utilize different privacy measures in the context of vertical federated learning to achieve effective feature aggregation. & The differential privacy may lead to performance degradation due to additional perturbations. \\
  \cline{3-5}
  \multicolumn{2}{|c|}{~}  & \makecell[c]{\cite{liu2020federated}} & Construct the forest algorithm using CART tree and bagging for both classification and regression tasks in vertical federated learning. & The privacy issues related to sharing intermediate embeddings and the potential privacy leakage of task labels have not been thoroughly explored.\\
  \cline{3-5}
  \multicolumn{2}{|c|}{~}  & \makecell[c]{\cite{yao2022efficient}} & Explore an efficient and robust vertical federated random forest algorithm for real-world large-scale data applications. & While data scalability is considered, the scale of the participants is also critical for efficiency.\\
  \cline{3-5}
  \multicolumn{2}{|c|}{~}  & \makecell[c]{\cite{hou2021verifiable}} & Propose a verifiable vertical federated random forest algorithm applicable to dynamic client scenarios. & There are additional communication costs associated with third-party homomorphic encryption.\\
  \cline{3-5}
  \multicolumn{2}{|c|}{~}  & \makecell[c]{\cite{lu2023squirrel}} & Squirrel integrates innovative GBDT algorithm designs with advanced cryptography, including methods to obscure sample distribution and optimize gradient aggregation efficiently. & The complexities involved in preserving privacy during training and the risks of publishing decision trees between clients have yet to be fully addressed.\\
  \cline{3-5}
  \multicolumn{2}{|c|}{\multirow{-23}{*}{\makecell[c]{Tree\\-based \\ Model}}} & \makecell[c]{\cite{li2023effective}} & Propose to integrate client-specific knowledge by adding layers to the tree structure, enabling accelerated training and scalability. & The method is particularly suited for tabular data and may not be directly applicable to other data types like images or text. \\
  \cline{1-5}
  \multicolumn{2}{|c|}{~}  & \makecell[c]{\cite{romanini2021pyvertical}} & PyVertical proposes a basic design utilizing split neural networks for vertical federated learning. & The privacy issue is severe as it directly exchanges intermediate features, yet lacks comprehensive discussions on privacy protection measures.\\
  \cline{3-5}
  \multicolumn{2}{|c|}{~}  & \makecell[c]{\cite{feng2020multi}} & MMVFL proposes to achieve label sharing through sparse learning and optimization with a global pseudo-label matrix. & Its applicability is limited to situations where all clients share the same label space.\\
  \cline{3-5}
  \multicolumn{2}{|c|}{~}  & \makecell[c]{\cite{liu2020secure}} & FTL achieves knowledge transfer through shared samples and feature space. & Its reliance on the shared feature space limits its application.\\
  \cline{3-5}
  \multicolumn{2}{|c|}{\multirow{-9}{*}{\makecell[c]{Neural \\ Network-\\based \\ Model}}} & \makecell[c]{\cite{abedi2024fedsl}} & FedSL constructs split Recurrent Neural Networks for handling sequential data in vertical federated learning. & It lacks discussions on scenarios where multiple data segments can be stored in a single client. \\
  \cline{1-5}
\end{tabular}
\end{table*}

\begin{table*}[t] 
\centering
\footnotesize
\caption{\textbf{Works on Feature \& Client Selection.}}
\label{tab:feature_client_selection}
\tabcolsep 2pt
\begin{tabular}{|c|c|c|m{5.7cm}|m{5.7cm}|} 
  \cline{1-5}
  \multicolumn{2}{|c|}{Topics} & \makecell[c]{Ref.} & \makecell[c]{Key Contributions} & \makecell[c]{Limitations} \\
  \cline{1-5}
  \multicolumn{2}{|c|}{~}  & \makecell[c]{\cite{li2023fedsdg}} & FedSDG-FS constructs dual-gate feature selection based on Gini impurity while ensuring privacy protection with differential privacy. & It assumes that all participants are honest, lacking a privacy discussion regarding malicious clients. \\
  \cline{3-5}
  \multicolumn{2}{|c|}{~}  & \makecell[c]{\cite{castiglia2023less}} & LESS-VFL proposes a communication-effective method based on the group lasso algorithm. & It introduces additional complexity with weight regularization terms. \\
  \cline{3-5}
  \multicolumn{2}{|c|}{~}  & \makecell[c]{\cite{feng2022vertical}} & It achieves feature selection through local feature construction and weight sparse learning. & The samples are not fully utilized for collaboration, which hinders the generalization of feature selection. \\
  \cline{3-5}
  \multicolumn{2}{|c|}{~}  & \makecell[c]{\cite{zhang2022secure}} & Build an efficient and privacy-preserving framework for eHealth data in vertical federated learning. & The features are selected only with the active client, while the remaining critical clients are ignored. \\
  \cline{3-5}
  \multicolumn{2}{|c|}{~}  & \makecell[c]{\cite{fu2023feast}} & FEAST constructs the feature selection based on the mutual information. & The greedy search selection may fail to converge to the optimal feature set. \\
  \cline{3-5}
  \multicolumn{2}{|c|}{\multirow{-12}{*}{\makecell[c]{Feature \\ Selection}}} & \makecell[c]{\cite{zhang2023embedded}} & PSO-EVFFS integrates evolutionary feature selection into the SecureBoost framework, optimizing both hyperparameters and feature subsets. & Its generalization to other backbones has not been explored. \\
  \cline{1-5}
  \multicolumn{2}{|c|}{~}  & \makecell[c]{\cite{wang2019measure}} & Construct a secure and fair contribution measurement based on Shapley value. & Entail assumptions that all clients hold the data with the same scale and have no correlations. \\
  \cline{3-5}
  \multicolumn{2}{|c|}{~}  & \makecell[c]{\cite{fan2022fair}} & The proposed VerFedSV achieves fair and efficient client contribution calculation by leveraging Shapley value principles. It is efficient to compute and retains many fairness properties.  & It relies on the preset timestamps which hinders its accuracy and stability. \\
  \cline{3-5}
  \multicolumn{2}{|c|}{\multirow{-7}{*}{\makecell[c]{Client \\ Selection \\
  }}} & \makecell[c]{\cite{jiang2022vf}} & VF-MINE calculates client contributions based on mutual information of each client. & The importance of participants may change over time, which client selection might not account for; The mutual information estimator will deviate to the small feature set, which will introduce bias for the final contribution estimation. \\
  \cline{1-5}
\end{tabular}
\end{table*}

\begin{table}[!t] 
\centering
\footnotesize
\caption{\textbf{Works on Addressing Privacy Leakage.}}
\label{tab:privacy_leakage}
\tabcolsep 2pt
\begin{tabular}{|c|c|c|m{5.25cm}|m{5.25cm}|}
  \cline{1-5}
  \multicolumn{2}{|c|}{Topics} & \makecell[c]{Ref.} & \makecell[c]{Key Contributions} & \makecell[c]{Limitations} \\
  \cline{1-5}
  \multicolumn{2}{|c|}{~}  & \makecell[c]{\cite{lu2020multi}} & Propose a multi-party private set intersection protocol that can handle cases where some parties drop out during the protocol execution. & It requires synchronous warm-up training at the beginning, and the gradient prediction method may introduce estimation errors. \\
  \cline{3-5}
  \multicolumn{2}{|c|}{~}  & \makecell[c]{\cite{sun2021vertical}} & FLORIST proposes to utilize a Private Set Union (PSU) protocol to align data entities without revealing sensitive membership information. & The PSU approach may increase the training sample size, leading to a potential rise in training costs. \\
  \cline{3-5}
  \multicolumn{2}{|c|}{\multirow{-8}{*}{\makecell[c]{Secure \\ Alignment
  }}}  & \makecell[c]{\cite{hardy2017private}} & Propose Privacy-Preserving Entity Resolution, where the raw data in clients is encrypted into a cryptographic long-term key (CLK). & The method may not be effective if identifiers are unstable or recorded with errors. \\
  \cline{1-5}
  \multicolumn{2}{|c|}{~}  & \makecell[c]{\cite{lu2020multi}} & Propose to leverage homomorphic encryption for intermediate embeddings to ensure data privacy without leaking information from raw data. & It introduces additional communication costs with a third-party coordinator for decryption and encryption. \\
  \cline{3-5}
  \multicolumn{2}{|c|}{~}  & \makecell[c]{\cite{wu2023falcon}} & Falcon supports VFL training for various backbones with strong privacy protection, using a hybrid strategy of threshold partially homomorphic encryption (PHE) and additive secret sharing scheme (SSS). & The need for threshold decryption in PHE and SSS conversion can be computationally expensive. \\
  \cline{3-5}
  \multicolumn{2}{|c|}{\multirow{-9}{*}{\makecell[c]{Secure \\ Embedding \\ Transportation
  }}} & \makecell[c]{\cite{wang2020hybrid}} & HDP-VFL enables joint learning of a generalized linear model from vertically partitioned data with minimal cost in terms of training time and accuracy. It leverages differential privacy (DP) to protect intermediate results during VFL, avoiding the need for Homomorphic Encryption or Secure Multi-Party Computation. & The accuracy of the joint model may be decreased with tight privacy budgets. \\
  \cline{1-5}
  \multicolumn{2}{|c|}{~}  & \makecell[c]{\cite{xu2021fedv}} & FedV enables collaborative training of machine learning models without peer-to-peer communication among parties. It uses functional encryption to secure gradient computation for models like linear regression, logistic regression, and SVMs. & The use of functional encryption may introduce computational complexity. \\
  \cline{3-5}
  \multicolumn{2}{|c|}{~}  & \makecell[c]{\cite{zhao2022vertically}} & CRDP-FL integrates differential privacy into a VFL framework to protect privacy while maintaining utility. By injecting differential privacy noise, CRDP-FL ensures strong privacy guarantees. & The trade-off between the performance and the privacy needs to be carefully considered. \\
  \cline{3-5}
  \multicolumn{2}{|c|}{\multirow{-12}{*}{\makecell[c]{Secure \\ Gradient \\ Transportation
  }}} & \makecell[c]{\cite{errounda2023adaptive}} &  AdaVFL adapts privacy protection to feature contributions and model convergence in vertical federated learning. It uses zero-Concentrated Differential Privacy (zCDP) for privacy accounting, aiming to balance privacy protection and utility. & It assumes honest-but-curious adversaries and secure communication protocol, which may not hold in practical scenarios. Besides, the adaptive budgeting scheme relies on the model convergence dynamics, which may vary with different datasets and learning tasks. \\
  \cline{1-5}
\end{tabular}
\end{table}

\begin{table}[!t] 
\centering
\footnotesize
\caption{\textbf{Works on Inference Attack.}}
\label{tab:inference_attack}
\tabcolsep 2pt
\begin{tabular}{|c|c|c|m{5.5cm}|m{5.5cm}|}
  \cline{1-5}
  \multicolumn{2}{|c|}{Topics} & \makecell[c]{Ref.} & \makecell[c]{Key Contributions} & \makecell[c]{Limitations} \\
  \cline{1-5}
  \multicolumn{2}{|c|}{~}  & \makecell[c]{\cite{luo2021feature}} & Constructs specific attacks with both single and multiple prediction outputs, with precise inference and generative inference, without any background information about the target data distribution. & The effectiveness of the attacks depends on the correlation between the features of the adversary and passive clients. \\
  \cline{3-5}
  \multicolumn{2}{|c|}{~}  & \makecell[c]{\cite{jin2021cafe}} & CAFE is designed to recover data in batches from shared aggregated gradients by leveraging data index and internal representation alignments. & The effectiveness of CAFE is influenced by the model architecture and weight distribution initialization. \\
  \cline{3-5}
  \multicolumn{2}{|c|}{~}  & \makecell[c]{\cite{rassouli2022privacy}} & Propose to utilize the Chebyshev center concept for attacks and provide theoretical performance guarantees. & Computing the Chebyshev center is complex and computationally intensive. \\
  \cline{3-5}
  \multicolumn{2}{|c|}{~}  & \makecell[c]{\cite{ye2022feature}} & It proves that while reconstructing general features is NP-hard, binary feature reconstruction is feasible and presents a search-based attack algorithm. & The attack primarily targets binary features. It may not be directly applicable to other types of data or features. \\
  \cline{3-5}
  \multicolumn{2}{|c|}{\multirow{-15}{*}{\makecell[c]{Feature \\ Inference \\ Attack
  }}} & \makecell[c]{\cite{yang2023practical}} & Propose utilizing an inference model to minimize the distance between predictions from inferred and target features. It employs zeroth-order gradient estimation to train the inference model without direct access to the global model and other local models. & The effectiveness of the attack is influenced by the correlation between known features and target features. Higher correlations lead to better attack performance.\\
  \cline{1-5}
  \multicolumn{2}{|c|}{~}  & \makecell[c]{\cite{liu2021batch}} & It introduces a gradient inversion model to reconstruct private labels with high accuracy by exploiting batch-averaged local gradients. It also presents a gradient-replacement attack that allows label replacement in black-boxed VFL without altering VFL protocols. & The success of the attack depends on the batch size being smaller than the dimension of the final fully connected layer. \\
  \cline{3-5}
  \multicolumn{2}{|c|}{~}  & \makecell[c]{\cite{li2021label}} & Propose a norm-based scoring function, which exploits the observation that the norm of the gradient vector potentially reveals label information. It uses the gradient norm as a predictor of the unseen label. & It is tailored to a specific threat model within two-party split learning, which may not generalize to other models or settings. \\
  \cline{3-5}
  \multicolumn{2}{|c|}{~} & \makecell[c]{\cite{fu2022label}} & Introduce label inference attacks exploiting the local model structure and gradient update mechanism in VFL. It proposes passive attacks using model completion, active attacks with a malicious local optimizer, and direct attacks by analyzing gradient signs. & The direct label inference attack is limited to training examples and requires additional steps for inference on new samples. \\
  \cline{3-5}
  \multicolumn{2}{|c|}{~}  & \makecell[c]{\cite{sun2022label}} & Introduce a label inference attack method that exploits the correlation between intermediate embeddings and private labels to steal sensitive label information. & The label party may need additional computational costs to re-learn the correlation between embeddings and labels to maintain model utility. \\
  \cline{3-5}
  \multicolumn{2}{|c|}{~}  & \makecell[c]{\cite{kariyappa2023exploit}} & Exploit frames the attack as a supervised learning task using gradient information obtained during split learning. & The efficacy of Exploit varies based on the chosen splitting layer. \\
  \cline{3-5}
  \multicolumn{2}{|c|}{\multirow{-25}{*}{\makecell[c]{Label \\ Inference \\ Attack
  }}} & \makecell[c]{\cite{qiu2022your}} & Propose a numerical approximation method designed to approximate encrypted representations, enabling the inference of private label-related relations. & The effectiveness of the attack is contingent on the knowledge of the adversary such as prediction results and global model parameters. \\
  \cline{1-5}
\end{tabular}
\end{table}

\begin{table}[!t] 
\centering
\footnotesize
\caption{\textbf{Works on Destructive Attack.}}
\label{tab:destructive_attack}
\tabcolsep 2pt
\begin{tabular}{|c|c|c|m{6.5cm}|m{4.2cm}|}
  \cline{1-5}
  \multicolumn{2}{|c|}{Topics} & \makecell[c]{Ref.} & \makecell[c]{Key Contributions} & \makecell[c]{Limitations} \\
  \cline{1-5}
  \multicolumn{2}{|c|}{~}  & \makecell[c]{\cite{liu2020backdoor}} & Propose recording the intermediate gradient of a clean sample with the targeted label and using this recorded gradient for poisoned samples. & The modification of gradients will influence the performance on the clean data. \\
  \cline{3-5}
  \multicolumn{2}{|c|}{~}  & \makecell[c]{\cite{chen2024universal}} & UAB utilizes bi-level optimization to optimize universal backdoor trigger generation and model parameters during VFL training, without additional data from other clients. & The performance will be affected if the dataset is imbalanced. \\
  \cline{3-5}
  \multicolumn{2}{|c|}{~}  & \makecell[c]{\cite{gu2023lr}} & Involve an adversarial participant in VFL who fine-tunes its local model to output specific latent representations for backdoor instances. It can be executed even without access to labels, using only local latent representations. & The attack performance may vary depending on the dataset and the number of classes. \\
  \cline{3-5}
  \multicolumn{2}{|c|}{~}  & \makecell[c]{\cite{he2023backdoor}} & Propose to inject a stealthy backdoor into the global model during training. It is achieved by replacing the local embeddings of a small number of target-class samples with a carefully constructed trigger vector, without modifying any labels. & The attack method is non-trivial and requires careful construction of the trigger vector. \\
  \cline{3-5}
  \multicolumn{2}{|c|}{~}  & \makecell[c]{\cite{bai2023villain}} & VILLAIN introduces a label inference algorithm to insert a backdoor into the global model. It enhances the backdoor attack power by designing a stealthy additive trigger and introducing backdoor augmentation strategies to impose a larger influence on the global model. & Its effectiveness relies on the sophisticated label inference algorithm, which introduces additional complexity. \\
  \cline{3-5}
  \multicolumn{2}{|c|}{~}  & \makecell[c]{\cite{xuan2023practical}} & BadVFL uses a Source Data Detection (SDD) module to trace data categories based on gradients and a Source Data Perturbation (SDP) scheme to enhance the decision dependency between the trigger and attack target. & The effectiveness of the attack may vary with the complexity of the global model structure, affecting gradient-based calculations. \\
  \cline{3-5}
  \multicolumn{2}{|c|}{\multirow{-23.5}{*}{\makecell[c]{Backdoor \\ Attack
  }}} & \makecell[c]{\cite{chen2022graph}} & Graph-Fraudster exploits the vulnerability of VFL for graph data by generating adversarial perturbations. It leverages noise-added global node embeddings and gradients of pairwise nodes to confuse the global model. & The success of the attack relies on the leakage of global node embeddings. \\
  \cline{1-5}
  \multicolumn{2}{|c|}{~}  & \makecell[c]{\cite{lai2023vfedad}} & VFedAD proposes three kinds of data-level poison attack methods: Random Failure, which adds noises to raw data for performing attacks; Random Mismatch, where the attack may shuffle the IDs in clients to disrupt the collaboration; Targeted Tampering, where the attacker attempts to replace some samples with target features. & It is easy to detect due to the obvious raw data modification. \\
  \cline{3-5}
  \multicolumn{2}{|c|}{~}  & \makecell[c]{\cite{qiu2024hijack}} & Introduce two attacks, the replay attack and the generation attack, to demonstrate the vulnerability of VFL systems to the Byzantine Generals Problem. & The success of attacks is affected by the completeness of features controlled by the adversary. Incomplete features can make attacks more challenging. \\
  \cline{3-5}
  \multicolumn{2}{|c|}{~}  & \makecell[c]{\cite{duanyi2023constructing}} & Propose to disrupt the inference process by adaptively corrupting a subset of clients. It formulates finding optimal attack strategies as an online optimization problem, involving adversarial example generation and corruption pattern selection. & The ability to adjust corruption patterns is based on the effectiveness of previous attacks, which may not always lead to the discovery of the optimal pattern. \\
  \cline{3-5}
  \multicolumn{2}{|c|}{\multirow{-16.5}{*}{\makecell[c]{Poison \\ Attack
  }}} & \makecell[c]{\cite{chen2024gan}} & It utilizes semi-supervised learning to create a surrogate target model, then employs a GAN-based method to generate adversarial perturbations that degrade the model performance. & The effectiveness is influenced by various factors like the number of attacker features and the known labels. \\
  \cline{1-5}
\end{tabular}
\end{table}

\begin{table}[!t] 
\centering
\footnotesize
\caption{\textbf{Works on Defending Against Inference Attack.}}
\label{tab:defense_inference_attack}
\tabcolsep 2pt
\begin{tabular}{|c|c|c|m{6.6cm}|m{4.7cm}|}
  \cline{1-5}
  \multicolumn{2}{|c|}{Topics} & \makecell[c]{Ref.} & \makecell[c]{Key Contributions} & \makecell[c]{Limitations} \\
  \cline{1-5}
  \multicolumn{2}{|c|}{~}  & \makecell[c]{\cite{sun2021defending}} & Propose an adversarial training framework, which makes the collaboration model robust against reconstruct attacks from shared gradients. & The adversarial training process adds complexity and requires careful tuning to achieve the desired balance between privacy and performance. \\
  \cline{3-5}
  \multicolumn{2}{|c|}{~}  & \makecell[c]{\cite{zhu2024vulnerabilities}} & VFLDefender employs gradient obfuscation to reduce the correlation between model updates and training data, effectively preventing reconstruction attacks. & It impacts model utility and the balance between privacy and performance needs further exploration. \\
  \cline{3-5}
  \multicolumn{2}{|c|}{~}  & \makecell[c]{\cite{chang2024gradient}} & Propose selectively transmitting a portion of the gradient components to reduce the risk of data leakage while maintaining model training accuracy. & It requires additional processing of gradient components, which are computationally intensive. \\
  \cline{3-5}
  \multicolumn{2}{|c|}{~}  & \makecell[c]{\cite{mao2022secure}} & Implement a privacy budget allocation scheme to perturb information exchange between the passive and the active client, protecting against data reconstruction attacks. & The dynamic privacy budget allocation method requires careful estimation of parameter importance, introducing additional complexity. \\
  \cline{3-5}
  \multicolumn{2}{|c|}{\multirow{-15}{*}{\makecell[c]{Defend \\ Against \\ Feature \\ Inference \\ Attack
  }}} & \makecell[c]{\cite{gu2023fedpass}} & FedPass introduces an adaptive obfuscation mechanism that adjusts during the learning process to protect features and labels simultaneously. It embeds private passports in both passive and active party models, making it exponentially hard for attackers to infer features and nearly impossible to infer private labels. & The randomness in passport generation is crucial for data privacy, but the process and its security implications require careful consideration. \\ 
  \cline{1-5}
  \multicolumn{2}{|c|}{~}  & \makecell[c]{\cite{sun2022label}} & Propose an additional optimization goal at the label party to minimize the distance correlation, making it difficult for an adversary to infer private labels from the shared intermediate embedding. & The method is primarily evaluated in a binary classification setting, and its effectiveness in other scenarios or with different data distributions is not extensively discussed. \\
  \cline{3-5}
  \multicolumn{2}{|c|}{~}  & \makecell[c]{\cite{li2021label}} & Propose minimizing label leakage in split learning by optimizing noise perturbation structures. It strategically adds random noise to gradients to prevent adversaries from recovering private labels. & It assumes a Gaussian distribution for unperturbed gradients and can be computationally intensive with multiple optimizations. \\
  \cline{3-5}
  \multicolumn{2}{|c|}{~}  & \makecell[c]{\cite{zou2022defense}} & DCAE proposes to defend against label inference and replacement attacks by disguising true labels using autoencoder and entropy regularization with prior label knowledge. & The implementation of DCAE may add complexity to the VFL system. \\
  \cline{3-5}
  \multicolumn{2}{|c|}{~}  & \makecell[c]{\cite{fan2023flsg}} & FLSG defends against passive label inference attacks by generating gradients similar to the original ones using a Gaussian distribution. & Implementing FLSG increases the training time compared to the original VFL framework. \\
  \cline{3-5}
  \multicolumn{2}{|c|}{~}  & \makecell[c]{\cite{wang2023beyond}} & Introduce a shadow model to share gradients during training, disrupting the correlation between gradients and training data, which hinders attackers from inferring labels. & The new local model and the need for secret sharing during training add complexity and potential cost to the process. \\
  \cline{3-5}
  \multicolumn{2}{|c|}{~}  & \makecell[c]{\cite{fu2024proj}} & ProjPert addresses label leakage by formulating an optimization problem that minimizes the impact on model quality while satisfying a pre-set privacy guarantee. & The heuristic variant may not always provide the optimal solution but is designed to be close to it. \\
  \cline{3-5}
  \multicolumn{2}{|c|}{~}  & \makecell[c]{\cite{qiu2024hashvfl}} & HashVFL addresses the challenges of learnability, bit balance, and consistency in VFL by employing a sign function for binarization, batch normalization for bit balance, and predefined binary codes for consistency.  & The integration of hashing can lead to vanishing gradients during model training. \\
  \cline{3-5}
  \multicolumn{2}{|c|}{~}  & \makecell[c]{\cite{zheng2022making}} & Propose the use of potential energy loss (PELoss) to make the output distribution of the local model more complex, protecting against label leakage. & The method assumes the attacker uses supervised learning to fine-tune the local model and does not consider attacks based on unsupervised learning approaches. \\
  \cline{3-5}
  \multicolumn{2}{|c|}{~}  & \makecell[c]{\cite{yang2022differentially}} & TPSL adds noise to gradients and model updates during training to ensure differential privacy, focusing on protecting sensitive label information. & The method involves adding noise to gradients and model updates, which may affect the utility of the trained model. \\
  \cline{3-5}
  \multicolumn{2}{|c|}{\multirow{-33}{*}{\makecell[c]{Defend \\ Against \\ Label \\ Inference \\ Attack}}} & \makecell[c]{\cite{takahashi2023eliminating}} & Propose defense mechanisms to mitigate label leakage in tree-based VFL, by utilizing label differential privacy with post-processing and mutual information regularization. & It requires additional training and communications. \\
  \cline{1-5}
\end{tabular}
\end{table}

\begin{table}[!t] 
\centering
\footnotesize
\caption{\textbf{Works on Defending Against Destructive Attack.}}
\label{tab:defense_destructive_attack}
\tabcolsep 2pt 
\begin{tabular}{|c|c|c|m{5.4cm}|m{5.4cm}|}
  \cline{1-5}
  \multicolumn{2}{|c|}{Topics} & \makecell[c]{Ref.} & \makecell[c]{Key Contributions} & \makecell[c]{Limitations} \\
  \cline{1-5}
  \multicolumn{2}{|c|}{~}  & \makecell[c]{\cite{he2023backdoor}} & Propose strategies from three aspects: Statistical Filtering, which filters out abnormal local embeddings that deviate from natural outputs; Reverse Engineering Detection, identifies backdoored models by comparing reversed trigger vectors for different classes; Confusional Autoencoder, which maps original labels to fake labels to minimize classification probability differences. & Can not detect stealthy attacks with well-crafted trigger vectors; With low detection rates, indicating difficulty in identifying malicious models; Does not significantly hinder the proposed attack, as it does not rely on gradients or label inference. \\
  \cline{3-5}
  \multicolumn{2}{|c|}{~}  & \makecell[c]{\cite{lai2023vfedad}} & VFedAD utilizes information theory to detect anomalies in vertical federated learning. It employs contrastive learning and cross-client prediction tasks to learn data representations that help identify poisoned samples. & The success of anomaly detection hinges on the quality of learned data representations. \\
  \cline{3-5}
  \multicolumn{2}{|c|}{~}  & \makecell[c]{\cite{qiu2024hijack}} & Utilize two strategies to defend the destructive attack: Normalization, which is effective against black-box attacks like ZOO by transforming perturbed input, preventing gradient approximation; Dropout, which reduces model memorization of specific patterns and mitigates attacks. & The impact is limited when the attacker has a significant feature ratio, as the threat persists; High dropout probability can drastically reduce main task performance, making it impractical for collaborative learning. \\
  \cline{3-5}
  \multicolumn{2}{|c|}{~}  & \makecell[c]{\cite{chen2024gan}} & Focus on a server-side anomaly detection algorithm based on a deep auto-encoder (DAE) to combat data poisoning attacks. It is used to identify outliers in embeddings with reconstruction errors and filter out anomalous data. & The detection might struggle with high proportions of poisoned data. \\
  \cline{3-5}
  \multicolumn{2}{|c|}{\multirow{-24}{*}{\makecell[c]{Defend \\ Against \\ Destructive \\ Attack}}} & \makecell[c]{\cite{liu2021rvfr}} & RVFR operates by training individual feature extractors seperately, followed by a robust feature subspace recovery process, and feature purification based on the assumption that only a small fraction of agents are malicious. & The effectiveness of RVFR relies on certain assumptions, such as a low-rank feature subspace and a small fraction of malicious agents. \\
  \cline{1-5}
\end{tabular}
\end{table}

\begin{table}[!t] 
\centering
\footnotesize
\caption{\textbf{Works on Limited Data.}}
\label{tab:limited_data}
\tabcolsep 2pt 
\begin{tabular}{|c|c|c|m{5.5cm}|m{5.5cm}|}
  \cline{1-5}
  \multicolumn{2}{|c|}{Topics} & \makecell[c]{Ref.} & \makecell[c]{Key Contributions} & \makecell[c]{Limitations} \\
  \cline{1-5}
  \multicolumn{2}{|c|}{~}  & \makecell[c]{\cite{kang2022fedcvt}} & FedCVT generates missing features and assigns pseudo-labels for unaligned samples. & The feature generation introduces noise information, and the unaligned samples are not fully utilized; only those with high-confidence pseudo-labels are employed. \\
  \cline{3-5}
  \multicolumn{2}{|c|}{~}  & \makecell[c]{\cite{sun2023communication}} & Few-shot VFL introduces the semi-supervised paradigm, where each client updates with estimated labels to improve sample utilization. & Unaligned samples are not fully utilized due to the confidence threshold.  \\
  \cline{3-5}
  \multicolumn{2}{|c|}{\multirow{-7}{*}{\makecell[c]{Limited \\ Aligned \\ Samples}}} & \makecell[c]{\cite{yang2022multi}} & FedMC expands the training data by matching non-overlapping samples based on similarity, thus improving the effectiveness of the jointly trained model. & It aims to pair the unaligned samples above the desired similarity threshold for collaboration training, which cannot fully utilize all samples. \\
  \cline{1-5}
  \multicolumn{2}{|c|}{~}  & \makecell[c]{\cite{cha2021implementing}} & They propose to leverage auto-encoder which does not require labels for training. & Redundant information is introduced and may overfit short sequence data.\\
  \cline{3-5}
  \multicolumn{2}{|c|}{~}  &  \makecell[c]{\cite{wu2022practical}} & FedOnce enhances intermediate features through local unsupervised learning. & It requires a sufficient amount of labels for the collaboration task; Additional computational costs are introduced with local updation.\\
  \cline{3-5}
  \multicolumn{2}{|c|}{~}  &  \makecell[c]{\cite{he2024hybrid}} & FedHSSL leverages cross-party views, local views, and invariant features of samples to improve the performance of VFL collaboration with limited labels. & Assume the participants have similar data distributions; Additional communication costs are introduced.\\
  \cline{3-5}
  \multicolumn{2}{|c|}{\multirow{-10}{*}{\makecell[c]{Limited \\ Labels}}} & \makecell[c]{\cite{castiglia2022self}} & SS-VFL leverages unlabeled data to train representation networks and labeled data for a downstream prediction network, aiming to achieve higher accuracy with reduced communication costs. & The effectiveness may vary depending on the similarity between local and centralized class probability distributions. Additionally, there is a potential for model bias if the datasets used for training are biased.  \\
  \cline{1-5}
\end{tabular}
\end{table}

\begin{table}[!t] 
\centering
\footnotesize
\caption{\textbf{Works on Communication Efficiency.}}
\label{tab:communication_efficiency}
\tabcolsep 2pt 
\begin{tabular}{|c|c|c|m{4.8cm}|m{4.8cm}|}
  \cline{1-5}
  \multicolumn{2}{|c|}{Topics} & \makecell[c]{Ref.} & \makecell[c]{Key Contributions} & \makecell[c]{Limitations} \\
  \cline{1-5}
  \multicolumn{2}{|c|}{~}  & \makecell[c]{\cite{zhang2021asysqn}} & AsySQN utilizes approximate second-order information to reduce the communication rounds and improve the convergence speed. & Require collaboration labels to be held by all clients. \\
  \cline{3-5}
  \multicolumn{2}{|c|}{~}  &  \makecell[c]{\cite{khan2022communication}} & They propose to utilize feature compression methods to compress the local data into latent representations and reduce the communication rounds to one. & The feature compression method should be carefully selected for different scenarios. \\
  \cline{3-5}
  \multicolumn{2}{|c|}{~}  & \makecell[c]{\cite{liu2022fedbcd}} & FedBCD utilizes a Federated Stochastic Block Coordinate Descent algorithm, enabling parties to perform multiple local updates to reduce communication rounds. & The number of local rounds needs careful selection to avoid divergence. \\
  \cline{3-5}
  \multicolumn{2}{|c|}{~}  &  \makecell[c]{\cite{castiglia2022compressed}} & They propose to leverage local updating and sharing compressed embeddings to reduce communication. & It is limited to the bounded embedding gradients and bounded Hessian of the objective function. \\
  \cline{3-5}
  \multicolumn{2}{|c|}{~}  &  \makecell[c]{\cite{fu2022towards}} & CELU-VFL utilizes cached stale statistics to estimate model gradients for local updating, thereby reducing frequent communications. & It involves approximation with stale statistics, which introduces potential errors in estimation. \\
  \cline{3-5}
  \multicolumn{2}{|c|}{~}  &  \makecell[c]{\cite{inoue2023sparsevfl}} & SparseVFL reduces the data size exchanged between the server and clients by exploiting the sparsity of embeddings and gradients. & It is not effective for dense embeddings/gradients or high-dimensional embeddings/gradients. \\
  \cline{3-5}
  \multicolumn{2}{|c|}{~}  &  \makecell[c]{\cite{wang2024unified}} & VFL-CZOFO applies a cascaded hybrid optimization to alleviate the slow convergence problem of ZO-based VFL, significantly reducing the communication cost. & It increases the computational cost for the server. Besides, it requires a trade-off between compression rate and test accuracy. \\
  \cline{3-5}
  \multicolumn{2}{|c|}{~}  &  \makecell[c]{\cite{huang2022cheetah}} & Cheetah introduces homomorphic encryption-based protocols for linear layers without expensive rotation operations and efficient primitives for non-linear functions, resulting in faster and more communication-efficient 2PC-NN inference. & It involves multiple computationally intensive steps, including partitioning input shares, encoding to polynomials, homomorphic operations, and re-masking.\\
  \cline{3-5}
  \multicolumn{2}{|c|}{\multirow{-33}{*}{\makecell[c]{Large \\ Communication \\ Burden}}} & \makecell[c]{\cite{gu2020federated}} & FDSKL uses random features to approximate kernel mapping functions and doubly stochastic gradients for solution updates, shown to be faster and more communication-efficient while maintaining similar generalization performance. & The model coefficients are stored separately across different clients to maintain privacy, which introduces additional complexity. \\
  \cline{1-5}
\end{tabular}
\end{table}

\begin{table}[!t] 
\centering
\footnotesize
\caption{\textbf{Works on Client Asynchrony.}}
\label{tab:client_asynchrony}
\tabcolsep 2pt 
\begin{tabular}{|c|c|c|m{5.1cm}|m{5.1cm}|}
  \cline{1-5}
  \multicolumn{2}{|c|}{Topics} & \makecell[c]{Ref.} & \makecell[c]{Key Contributions} & \makecell[c]{Limitations} \\
  \cline{1-5}
  \multicolumn{2}{|c|}{~}  & \makecell[c]{\cite{chen2020vafl}} & They propose utilizing stochastic gradient algorithms independently on each client without coordination, allowing clients to participate intermittently or strategically. & It may become complex when dealing with nonlinear local embedding functions. \\
  \cline{3-5}
  \multicolumn{2}{|c|}{~}  & \makecell[c]{\cite{li2020efficient}} & They propose an efficient method for handling client asynchrony, by utilizing gradient prediction through Taylor expansion and double-end sparse compression to reduce training costs. & It requires synchronous warm-up training at the outset, and the gradient prediction method may introduce estimation errors. \\
  \cline{3-5}
  \multicolumn{2}{|c|}{~}  & \makecell[c]{\cite{gu2021privacy}} & AFSGD-VP enhances efficiency by ensuring all computational resources remain engaged, eliminating idle time waiting for synchronization. & As the number of clients grows, the communication overheads can restrict the scalability and speedup of the algorithms. \\
  \cline{3-5}
  \multicolumn{2}{|c|}{~}  & \makecell[c]{\cite{zhang2021secure}} & They propose a bilevel asynchronous parallel architecture, enabling asynchronous model updates across parties. & It assumes that one or partial parties hold labels, which may not be the case in practical applications. \\
  \cline{3-5}
  \multicolumn{2}{|c|}{~}  & \makecell[c]{\cite{shi2022practical}} & They propose leveraging asynchronous training to reduce waiting time. & The computational complexity and communication cost may increase with the number of participants and the data size. \\
  \cline{3-5}
  \multicolumn{2}{|c|}{~}  & \makecell[c]{\cite{qiu2023vfedsec}} & Utilize a robust aggregation mechanism capable of handling missing or temporarily disconnected clients. & It might require significant computational costs, rendering it less practical for real-time applications. \\
  \cline{3-5}
  \multicolumn{2}{|c|}{~}  & \makecell[c]{\cite{sun2023robust}} & Propose party-wise dropout to prevent the co-adaptation of feature extractors across parties. & The model may rely on certain representation components from specific parties. \\
  \cline{3-5}
  \multicolumn{2}{|c|}{\multirow{-23}{*}{\makecell[c]{Client \\ Asynchrony
  }}}  & \makecell[c]{\cite{li2023fedvs}} & It uses secret sharing schemes for local data and models, and the aggregation of embeddings is reconstructed from non-straggling clients. & The reconstruction may have limitations when the number of non-straggling clients is limited. \\
  \cline{1-5}
\end{tabular}
\end{table}
\end{document}